\documentclass[manuscript,screen]{acmart}
\usepackage[utf8]{inputenc}
\usepackage{textcomp}
\usepackage[acronym]{glossaries}
\usepackage[english]{babel}
\usepackage{amsthm}
\usepackage{graphicx}
\usepackage{subcaption}
\usepackage{hyperref}
\usepackage{amsmath}
\usepackage{multirow}
\usepackage{pdflscape}
\usepackage{import}
\usepackage{afterpage}
\usepackage{array}
\usepackage[export]{adjustbox}
\usepackage{tabularray}
\usepackage[inkscapearea=page]{svg}
\usepackage{makecell}
\usepackage{float}

\usepackage{longtable}
\graphicspath{ {./images/} }

\theoremstyle{definition}
\newtheorem{definition}{Definition}

\theoremstyle{remark}

\newacronym{ai}{AI}{Artificial Intelligence}
\newacronym{cv}{CV}{Computer Vision}
\newacronym{nlp}{NLP}{Natural Language Procesing}
\newacronym{dl}{DL}{Deep Learning}
\newacronym{ml}{ML}{Machine Learning}
\newacronym{dnn}{DNN}{Deep Neural Networks}
\newacronym{cnn}{CNN}{Convolutional Neural Networks}

\newacronym{iou}{IoU}{Intersection over Union}
\newacronym{miou}{mIoU}{mean Intersection over Union}
\newacronym{fbiou}{FB-IoU}{Foreground Background Intersection over Union}
\newacronym{map}{MAP}{Masked Averge Pooling}
\newacronym{fid}{FID}{Frécht Inception Distance}

\newacronym{fsl}{FSL}{Few Shot Learning}
\newacronym{fsc}{FSC}{Few Shot Classification}
\newacronym{fsod}{FSOD}{Few Shot Object Detection}
\newacronym{fss}{FSS}{Few Shot Semantic Segmentation}
\newacronym{ifss}{iFSS}{Incremental Few Shot Segmentation}
\newacronym{GFS-Seg}{GFS-Seg}{Generalized Few-shot Semantic Segmentation}
\newacronym{cdfss}{CF-FSS}{Cross-Domain Few-Shot Semantic Segmentation}
\newacronym{nway_kshot_fss}{N-Way K-Shot SS}{N-Way K-Shot Semantic Segmentation}

\newacronym{GAN}{GAN}{Generative Adversarial Network}
\newacronym{VAE}{VAE}{Variational Auto Encoder}

\AtBeginDocument{%
  \providecommand\BibTeX{{%
    \normalfont B\kern-0.5em{\scshape i\kern-0.25em b}\kern-0.8em\TeX}}}

\setcopyright{acmcopyright}
\copyrightyear{2018}
\acmYear{2018}
\acmDOI{XXXXXXX.XXXXXXX}

\acmJournal{JACM}
\acmVolume{37}
\acmNumber{4}
\acmArticle{111}
\acmMonth{8}

\begin{document}

\title[FSS: a review of methodologies, benchmarks, and open challenges]{Few Shot Semantic Segmentation: a review of methodologies, benchmarks, and open challenges}

\author{Nico Catalano}
\email{nico.catalano@polimi.it}
\orcid{0009-0004-2731-1068}

\author{Matteo Matteucci}
\email{matteo.matteucci@polimi.it}
\orcid{0000-0002-8306-6739}
\affiliation{%
  \institution{Politecnico di Milano}
  \streetaddress{Piazza Leonardo DaVinci 32}
  \city{Milano}
  \country{Italy}
  \postcode{20133}
}


\begin{abstract}
Semantic segmentation, vital for applications ranging from autonomous driving to robotics, faces significant challenges in domains where collecting large annotated datasets is difficult or prohibitively expensive. In such contexts, such as medicine and agriculture, the scarcity of training images hampers progress.
Introducing Few-Shot Semantic Segmentation, a novel task in computer vision, which aims at designing models capable of segmenting new semantic classes with only a few examples. This paper consists of a comprehensive survey of Few-Shot Semantic Segmentation, tracing its evolution and exploring various model designs, from the more popular conditional and prototypical networks to the more niche latent space optimization methods, presenting also the new opportunities offered by recent foundational models.
Through a chronological narrative, we dissect influential trends and methodologies, providing insights into their strengths and limitations. A temporal timeline offers a visual roadmap, marking key milestones in the field's progression.
Complemented by quantitative analyses on benchmark datasets and qualitative showcases of seminal works, this survey equips readers with a deep understanding of the topic. By elucidating current challenges, state-of-the-art models, and prospects, we aid researchers and practitioners in navigating the intricacies of Few-Shot Semantic Segmentation and provide ground for future development.

\end{abstract}

\ccsdesc[500]{Computing methodologies~Image segmentation}
\ccsdesc[300]{Computing methodologies~Machine learning}

\keywords{few shot learning, prototypical networks, conditional networks, latent space optimization, contrastive learning, generative models, variational auto encoders}

\received{20 February 2007}
\received[revised]{12 March 2009}
\received[accepted]{5 June 2009}

\maketitle

\section{Introduction}
\label{introduction}
\begin{figure}[t]
    \centering
    
    \begin{subfigure}[t]{.32\textwidth}
        \centering
        \includegraphics[width=1\linewidth]{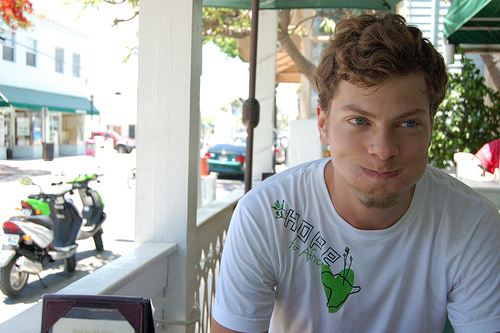}
        \caption{Original image }
        \label{fig:cv_tasks_original}
    \end{subfigure}
    \begin{subfigure}[t]{.32\textwidth}
        \centering
        \includesvg[width=1\linewidth]{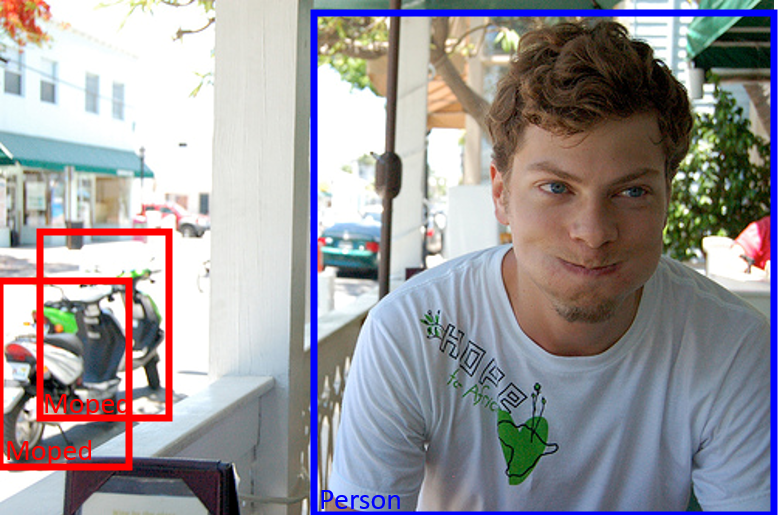}
        \caption{Object detection}
        \label{fig:cv_tasks_detection}
    \end{subfigure}
    \begin{subfigure}[t]{.32\textwidth}
        \centering
        \includegraphics[width=1\linewidth]{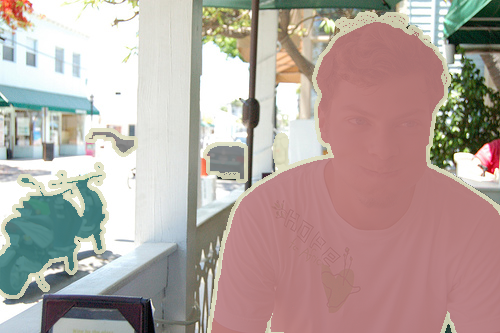}
        \caption{Semantic Segmentation}
        \label{fig:cv_tasks_semantic}
    \end{subfigure}
    \begin{subfigure}[t]{.33\textwidth}
        \centering
        \includegraphics[width=1\linewidth]{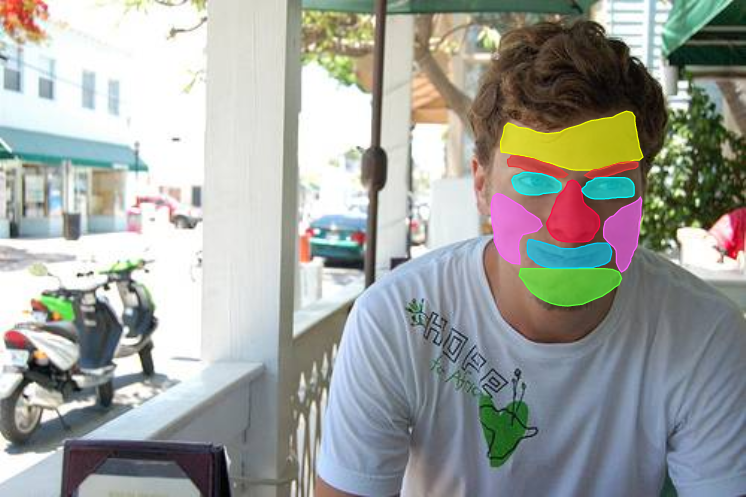}
        \caption{Parts Segmentation}
        \label{fig:cv_tasks_parts_segmentation}
    \end{subfigure}
    \begin{subfigure}[t]{.32\textwidth}
        \centering
        \includegraphics[width=1\linewidth]{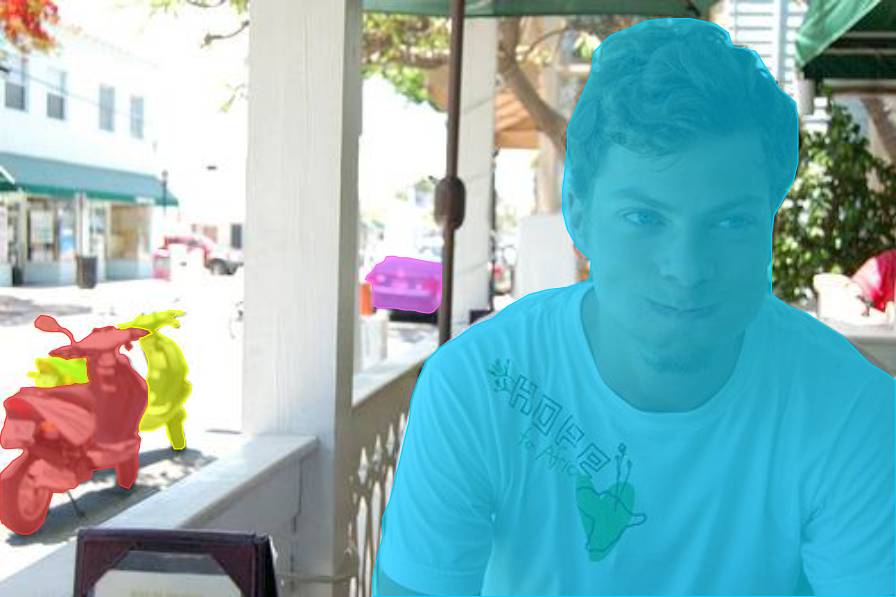}
        \caption{Instance Segmentation}
        \label{fig:cv_tasks_instance}
    \end{subfigure}
    \begin{subfigure}[t]{.32\textwidth}
        \centering
        \includesvg[width=1\linewidth]{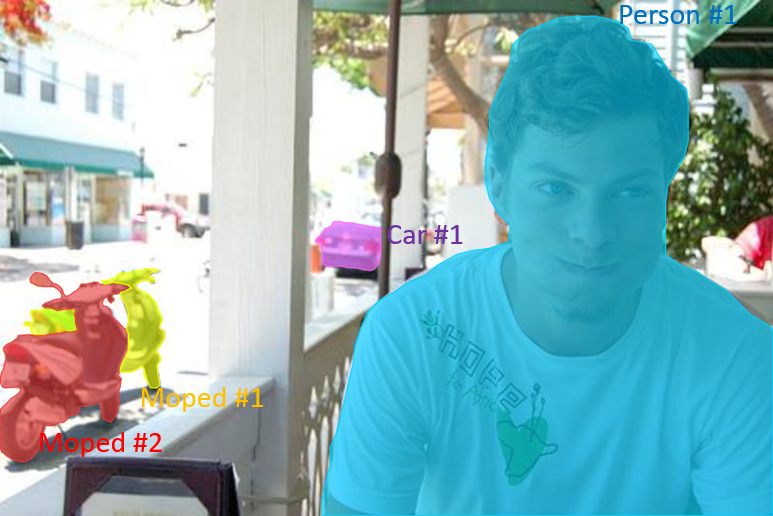}
        \caption{Panoptic Segmentation}
        \label{fig:cv_tasks_panoptic}
    \end{subfigure}

    \caption{Comparison of various \acrfull{cv} tasks. In a given image (depicted in picture a), there are multiple layers of interpretation possible. Object Detection (picture b) identifies the region containing a subject using bounding boxes. Semantic Segmentation (picture c) assigns a class label to each pixel without necessarily distinguishing different instances of the same subject class. Parts Segmentation (picture d) provides pixel-level annotation by segmenting the constituent parts of the intended subject, such as the face parts in this example. Instance Segmentation (picture e) distinguishes the subjects in the scene with a different segmentation mask for each one but does not necessarily assign a semantic class. Panoptic Segmentation (picture f) involves the semantic classification of every pixel in an image, coupled with the delineation of each unique subject instance mask. In the given image, two mopeds are identified with labels ``Moped\_1" and ``Moped\_2," illustrating the distinct instances within the same class.}
    \label{fig:cv_tasks}
\end{figure}

\acrfull{cv} plays a crucial role in various application fields, from robotics and clinical analysis to video surveillance; semantic segmentation represents one of the most significant tasks in this research field. Semantic segmentation can be described as predicting pixel-level category labels, thereby generating a segmentation mask for fixed subject categories in a given image \cite{garcia2018survey,guo2018review,8637467}. 
To better contextualize the role of semantic segmentation, it is essential to understand its relationships with other \acrshort{cv} tasks such as image classification, object detection, parts segmentation, etc. Image classification focuses on comprehending the overall scene by assigning one or more labels to an entire image (see Figure~\ref{fig:cv_tasks_original}). Object detection (see Figure \ref{fig:cv_tasks_detection}) concentrates on predicting object locations, typically supplying bounding boxes for the identified objects. Parts segmentation (depicted in Figure~\ref{fig:cv_tasks_parts_segmentation}) closely resembles semantic segmentation, aiming to predict pixel-level segmentation masks that cover distinct parts constituting the intended subject. Instance segmentation (see Figure \ref{fig:cv_tasks_instance}) aims at differentiating individual subjects in an image, even if they share the same kind, without necessarily assigning them a category. Finally, panoptic segmentation (Figure \ref{fig:cv_tasks_panoptic}) seamlessly combines semantic segmentation with instance segmentation, predicting pixel-level categories and distinguishing each class instance in the scene. Figure \ref{fig:cv_tasks}, which visually compares various CV tasks, puts semantic segmentation as a crucial intermediary task, as it provides enough image understanding to enable a large number of downstream applications but still lacks deeper analysis and interpretation.

The advent of \acrfull{dnn} and \acrfull{cnn} of recent years has revolutionized the \acrfull{ai} community and the \acrfull{cv} research field allowing the introdocution of semantic segmentation models capable of impressive accuracy, with works by Long et al. \cite{long2015fully}, Badrinarayanan et al. \cite{badrinarayanan2017segnet}, Ronneberger et al. \cite{ronneberger2015u}, Chen et al. \cite{chen2018encoder}, and Zhao et al. \cite{zhao2017pyramid} underscoring the potential of these architectures. However, a significant drawback emerges for semantic segmentation models: their reliance on extensive training datasets.

Creating and labeling large datasets for semantic segmentation can demand substantial resources. For example, labeling object instances for 80 classes in 328,000 images for the MS COCO \cite{MS-COCO} dataset required 55,000 worker hours. This intensive data collection process poses significant financial and temporal challenges, especially in domains like agriculture or healthcare, where data variability and scarcity, requirements of domain-specific knowledge, and privacy concerns hinder acquisition efforts. Indeed, the desire to develop models capable of generalizing from a limited number of training samples is not new to the \acrfull{ai} and \acrfull{ml} communities and falls under the definition of \acrfull{fsl}  \cite{review_zero_one_few_learning_Kadam,wang2020generalizing,huang2022survey,ren2022visual}. Further specilizing the \acrshort{fsl} problem to the semantic segmentation task takes the name of \acrfull{fss}, and it is the focus of our contribution. 

In particular, this work discusses the main approaches and strategies adopted to design models capable of learning to segment a new semantic class with just few labelled examples (e.g., from 1 to 5). Our examination navigates the main families of key solutions proposed in the \acrshort{fss} literature and describes commonly used public benchmark datasets that serve as crucial evaluation tools for \acrshort{fss} models. Additionally, we present and compare the outcomes of noteworthy works in the field, providing both quantitative results on benchmark datasets as well as selected samples of predictions, providing this way also qualitative results. Furthermore, building upon the insights gained from previous surveys \cite{chang2023few,ren2022visual}, we extend our discussion to encompass the outlooks and future directions of \acrshort{fss}. This includes an examination of the rapidly expanding and promising innovations introduced with vision foundational models. In our conclusion, we address primary critiques and potential extensions, offering a nuanced overview of the current landscape and illuminating potential future trajectories in this evolving field.

A key distinction of our work from previous surveys, such as that of Ren et al. \cite{ren2022visual}, lies in its focused scope. While Ren et al. explored the broader scope few and zero-shot visual semantic segmentation methods across 2D and 3D spaces, this survey is dedicated exclusively to  \acrshort{fss}, providing a more updated and detailed analysis specifically tailored to the segmentation task in the few-shot scenario. This focused approach allows for a deeper understanding and more comprehensive evaluation of techniques relevant to this particular area. Furthermore, our survey paper fills a critical gap in the literature by offering a specialized examination of \acrshort{fss}. Although previous surveys have touched upon related topics, such as \cite{chang2023few,ren2022visual}, the rapid advancements and unique challenges within the realm of few-shot segmentation necessitate dedicated attention. By focusing solely on this niche, we are able to provide nuanced insights, identify emerging trends, and highlight key areas for future research.

\section{Background and Problem Definition}

\acrfull{fss} integrates principles from \acrfull{fsl} into the domain of semantic segmentation by specializing the semantic segmentation task to a low data regime. To elucidate the shared concepts among \acrshort{fss} and \acrshort{fsl}, we commence by recalling the fundamental definition of \acrfull{ml}, as articulated by Mitchell et al. \cite{mitchell1997machine} and Mohri et al. \cite{mohri2018foundations}:

\begin{definition}[Machine Learning \cite{mitchell1997machine, mohri2018foundations}]
A computer program is said to learn from experience $E$ with respect to some classes of task $T$ and performance measure $P$ if its performance can improve with $E$ on $T$ measured by $P$.

\end{definition}

\noindent
This fundamental notion lays the groundwork for the concept of \acrfull{fsl}, where the learning paradigm is focused on scenarios where available experience $E$ is limited. Wang et al. \cite{wang2020generalizing} define this ideas as \acrshort{fsl}:

\begin{definition}[\acrfull{fsl} \cite{wang2020generalizing}]
\acrfull{fsl} is a type of machine learning problems (specified by $E$, $T$, and $P$), where $E$ contains only a limited number of examples with supervised information for the target $T$.

\end{definition}

This transition from the broader landscape of \acrshort{ml} to the nuanced realm of \acrshort{fsl} becomes particularly pertinent when applied to the challenges of semantic segmentation. The objective of this task is to predict a segmentation mask over an input image covering the intended class of subject. Models for this task often rely on large-scale training datasets, presenting a significant challenge in scenarios where data availability is limited. Here, the concept of \acrfull{fss} emerges as a natural progression, seamlessly integrating principles from \acrshort{fsl}, as articulated by Shaban et al. \cite{BMVC2017_167}:

\begin{definition}[\acrfull{fss} \cite{BMVC2017_167}]
\acrfull{fss} is the problem of predicting the semantic segmentation mask $\hat{M}_q$ of a subject class $l$ in a query image $I_q$ given a support set $S$ composed by a limited training set of $k$ image mask pairs.
\end{definition}

This definition can be further formalized operationally as it follows. Given a semantic class $l$, we construct a support set of $k$ image-mask pairs:
\[
    S(l) = \{(I^i, M^i_l)\}_{i=1}^k,
\]
where, $I^i$ represents an RGB image of shape $[H^i,W^i,3]$, and $M^i_l$  is a binary mask of shape  $[H^i,W^i]$ segmenting the class $l$ in $I^i$. Additionally, a query image $I_q$ is considered, and the objective is to learn a model $f_{\theta}$ such that:
\[
   \hat{M_q} = f_{\theta}(I_q,S(l)).
\]

This model predicts the binary mask $\hat{M}_q$ for the semantic class $l$ in the query image $I_q$. In this context, the number of available shots $k$ is crucial, leading to the designation of \acrshort{fss} as K-FSS in various works, with $k$ often set to 1 or 5 \cite{BMVC2017_167, rakelly2018conditional, zhang2020sg, zhang2019canet, tian2020prior, liu2020part, wu2021learning, iqbal2022msanet}.


Given the scarcity of labelled samples, traditional algorithms can be inadequate for training \acrshort{fss} models. In order to extract maximum learning from the limited number of examples, most studies \cite{BMVC2017_167,rakelly2018conditional,zhang2019canet,boudiaf2021few, tian2020prior,wu2021learning,nguyen2019feature,li2021adaptive,lu2021simpler,siam2019adaptive,wang2019panet,Dong2018FewShotSS} have employed episodic training which is a form of meta-learning, or learning-to-learn approach. Another common strategy is transfer learning, which capitalizes on the knowledge acquired by pre-training a portion of the model on different tasks or datasets. Being it a basic component, the following section presents an overview of the use of episodic training in \acrshort{fss} and it explains how pre-training can be applied to \acrshort{fss}.

\subsection{Episodic training} \label{episodic_training}

The main characteristic of \acrshort{fss} is its limited number of labelled examples. This aspect is particularly challenging for training procedures, rendering traditional methods inadequate. To better adapt to the \acrshort{fss} settings, it is common choice to resort to episodic training. Episodic training is a meta-learning methodology where the model is trained through a series of “episodes”. According to this approach, initially proposed by Vinyals et al. in \cite{matching_networks}, a model can be trained with a series of meta-learning episodes composed by a support set of labeled images $S$, a query image  $I_q$ and its corresponding ground truth mask  $M_q$. The model then learns by minimizing the loss between the predicted mask  $\hat{M}_q$ and the ground truth mask ${M}_q$ for each episode. The model can later be tested with meta-testing episodes on unseen classes. 

This concept can be detailed as it follows: given the set $C$ of the classes in the dataset $D_{train}$, we split it in  $C_{train}$ and $C_{test}$ such that $C_{train} \cap C_{test} = \emptyset$. 
We can now form a training episode by randomly sampling one label class $c$ from $C_{train}$. Fixed $c$, we sample $k$ unique images and their corresponding masks segmenting the class $c$ in them to form the support set $S$, and one image query $I_q$ and its ground truth ${M}_q$ for the class $c$. Given the support set $S$, the query image $I_q$ and the ground truth ${M}_q$ as described, the training objective of the model is:
\begin{equation} \label{eq:episodic_training}
\hat{\theta} =\arg \max _{\theta} \mathbb{E}_{S , {(I_q, M_q)} }\left[ \log P_{\theta}(M_q \mid I_q, S)\right] 
\end{equation}

Optimizing  $\theta$  in \autoref{eq:episodic_training} implements a form of meta-learning since, in each episode, the model is learning to learn how to predict a segmentation mask for an unseen class given only a limited support of examples. Thus, with enough episodes, the model is expected to learn how to generalize from a limited support set.

\subsection{Transfer Learning}\label{pretraining}
In order to maximize the information extraction from limited data, a prevalent strategy within the literature involves the implementation of transfer learning, a set of techniques in which knowledge acquired from one task or dataset is applied to another related domain. In the context of \acrshort{fss} one common implementation of transfer learning is the use of pretrained feature extractors or backbones. These feature extractors can be initially trained to learn general visual representations on large datasets, such as the object detection dataset ImageNet \cite{ImageNet}. Leveraging this idea, works such as those by \cite{BMVC2017_167, rakelly2018conditional, Dong2018FewShotSS, zhang2020sg, nguyen2019feature} integrate a frozen pretrained feature extractor into their pipelines allowing for the extraction of domain-agnostic visual cues, such as shapes and patterns.

While the reliance on a large dataset for pretraining may seem contradictory to the low-data regime of \acrshort{fss}, it is crucial to recognize that the term `few shots' pertains solely to the new class that the model needs to learn. Therefore, under the condition that the semantic classes utilized in pretraining are distinct from those to be learned with few shots, pretrained feature extractors are a permissible form of transfer learning in this context.

In sum, to fully exploit the few examples given for each new class, a common practice found in the literature is to perform pre-training on an external $D_{pretrain}$ like ImageNet and train on the given dataset $D_{train}$  using episodic learning. In this scheme, a model can have a pre-training dataset $D_{pretrain}$ and a training dataset $D_{train}$ where $D_{pretrain} \neq  D_{train}$. 
In the field of \acrshort{fss}, Episodic learning is implemented by splitting the label set of all classes  $C$  in the dataset $D_{train}$  in $C_{train}$ and $C_{test}$ such that $C_{train} \cap C_{test} = \emptyset$.
In addition, a portion of the model can be pre-trained on $D_{pretrain}$, and the entire model can undergo meta-train on episodes from $D_{train}$ with label class from $C_{train}$, and meta-test with episodes from $D_{test}$ on classes from $C_{test}$.

\section{Methodology Spectrum} 

The field of \acrshort{fss} can be a multifaceted subject that can be approached from various perspectives. This section encompasses some of the most influential and notable studies in this area. To accommodate a diverse range of works and research lines and avoid overly strict categorization, we start identify three main strategies: conditional networks,  prototypical learning, and latent space optimization. Latent space optimization can be further specialized as generating synthetic datasets, e.g., by using GAN encodings, and via contrastive learning. Moreover, in this section we present the new opportunities offered by the recent foundation models and how they can be applied to tackle the \acrshort{fss} task. \autoref{fig:taxonomy} depicts a timeline indicating the most influential \acrshort{fss} models and the definitions of the task. In addition, the proposed timeline reports the trends and influences of the approaches presented in this paper.
In the following sections, we describe each approach and highlight the key details of representative works.

    \begin{figure}[t]
        \centering
        \includesvg[inkscapelatex=false,width=\linewidth]{svg/timeline_papers_strategies.svg}

        \caption{Timeline illustrating the progression of the \acrshort{fss} research field. Some of the most consequential models are depicted at the top, with colored bands representing design trends and influences. Overlapping and fading bands suggest the sharing of concepts between multiple works to varying degrees. Below, key milestones mark the definition of the \acrshort{fss} problem and its variations.}
        \label{fig:taxonomy}
        
    \end{figure}
    
    \subsection{Conditional Networks}
        The earliest works in the field, by Shaban et al. \cite{BMVC2017_167} and Rakelly at al. \cite{rakelly2018conditional}, addressed the \acrshort{fss} challenge by using two branched models as depicted in \autoref{fig:conditioning}. This kind of models has two parallel branches: the conditioning branch and the segmentation branch. 
The conditioning branch $g$ takes in input the support set $S$ and produces a parameter set $\theta$. 
The segmentation branch $h$ gets as input the query image $I_q$ and the parameter set $\theta$ generated by the conditioning branch $g$ to produce the prediction mask $\hat{M}_q $. 
During testing, the model works by first extracting a dense feature volume from $I_q$ using a pre-trained feature extractor in the segmentation branch $h$. Then the segmentation branch $h$ utilizes both the feature volume and the parameter set $\theta$  to predict the final mask $\hat{M_q}$.
We can model the conditioning branch $g$  as a function which takes as input the support set $S$ and produces a set of parameters:
\[ \theta = g(S) \]
On the other hand, the segmentation branch $h$ extracts a feature volume from $I_q$ using an embedding function $\phi$. So let $F_q = \phi(I_q)$ be the feature volume, and $F^{mn}_q$ be the  feature vector at the spatial location $(m,n)$, the segmentation branch uses the parameters $\theta$ and the feature vector $F^{mn}_q$ to predict the mask at the spatial location $(m,n)$ as:
\[ \hat{M}^{mn}_q = h(F^{mn}_q, \theta). \]

Rakelly et al. in \cite{rakelly2018conditional} experimented with this setup by designing two different parameter sets $\theta$. In the first experiment, $\theta$ is the feature volume extracted from the support set, which is fused with the query features in the segmentation branch. The resulting feature volume is then decoded to a binary mask $\hat{M}_q$ by a small convolutional network $f_{\theta}$ that can be interpreted as a learned distance metric for retrieval from support to query.
In the second experiment, $\theta$ is a set of linear classifier weights to be applied to the query feature volumes.
The results of \cite{rakelly2018conditional} show that the first approach leads to better outcomes, while Shaban et al. in \cite{BMVC2017_167} achieved slightly better predictions by using the parameter set $\theta$  to implement pixel-level logistic regression on the feature volume.

Lu et al. \cite{lu2021simpler} proposed a similar overall architecture based on transformers \cite{vaswani2017attention} to predict $\theta$. Similar to previous works in this category, $\theta$ is a set of weights for a classifier in charge of generating the final mask, and the feature extractor is pre-trained and shared between the conditioning and segmentation branches. 
Here, the key is the transformer within the segmentation branch as it utilizes the linear classifier weights $\theta$ from the conditioning branch as the query, and the feature vectors as the key and value. The resulting final weights are then used on a classifier to predict the subject mask on the query image.

Reckoning that different parameter sets $\theta$ can help the segmentation on different key aspects on the query image, Zhang et al.  \cite{9463398} propose leveraging the diversity of three distinct parameter sets. They propose the computation of three distinct similarity maps using these parameter sets, which are subsequently fused to generate a final prediction.
The first parameter set, termed Peak Embedding, guides the segmentation process by identifying and emphasizing the most distinctive features in the query image. This is achieved by highlighting features with higher values relative to the subject embeddings, thereby directing attention towards salient image regions.
The second parameter set, known as Global Embedding, focuses on capturing the overarching patterns of the subject within the support image. By computing the average vector of features within the object region, this embedding encapsulates the global context required for accurate segmentation.
Lastly, the Adaptive Embedding is derived using an attention mechanism to discern the relative importance of object-related pixels via self-attention. This adaptive approach enables the model to dynamically adjust its focus based on the content and context of the input, enhancing segmentation performance in complex scenarios.

For these kinds of solutions, the choice of $g$ and $h$ is crucial, as the design of the conditioning branch and the interplay that its output $\theta$ has within the segmentation branch can significantly influence predictions. Specifically, the relation between the parameter set $\theta$, the query features, and the expected prediction might be nontrivial and, therefore, challenging to learn for the segmentation branch. 
Conditional networks were the earliest model structure employed for the \acrshort{fss} task, and the shortcomings deriving from the use of two distinct functions for the support set and the query image have been mitigated with the introduction of the prototypical networks described in the next section. 

\begin{figure}[t]
    \centering
    \includesvg[width=\linewidth]{svg/new_conditioning_nets.svg}
    
    \caption{Conditional networks adapted from \cite{rakelly2018conditional}. Using the support set $S(l)$, the conditioning branch produces the parameter set $\theta$. The segmentation branch then uses $\theta$ to predict a mask $\hat{M}_q$ over the query image $I_q$.
    }
    \label{fig:conditioning}
    
\end{figure}
    
    \subsection{Prototypical Networks}  
       \begin{figure}[t]
    \centering
    \includesvg[width=0.7\linewidth]{svg/new_snell_prototypes-2}
        
        
    \caption{Prototypical networks adapted from \cite{snell2017prototypical}. The prototypes are computed as the mean $c_k$ of the embeddings from the same class in the support set identified in the picture with distinct colors.  The label for the query embedding $X$  is then assigned based on which prototype it is closer.
    }
    \label{fig:prototype}
    
\end{figure}


The challenges of \acrshort{fss} can be also approached with metric learning as reference framework. Prototypical Networks, which were initially introduced for  \acrfull{fsc} by Snell et al. \cite{snell2017prototypical}, have been adapted for \acrshort{fss} by Dong et al. \cite{Dong2018FewShotSS} and have become a popular strategy in many works in the field \cite{wang2019panet,zhang2020sg,liu2020part, zhang2020sg, zhang2019canet,Li2021FewshotSS}.
Prototypical networks are based on the intuition that images can be represented by points in some embedding space where images of similar objects are represented by points close together, while points that are far away represent images of different objects. Therefore, computing the centroid of all the points associated with images of the same class in the support set will give you a point representing a prototype for that class. At inference time, the label of a new image can be predicted by looking at the nearest prototype. A graphical intuition of this concept is proposed in \autoref{fig:prototype}.
Given that segmentation can be modelled as pixel-wise classification,  Dong et al. in \cite{Dong2018FewShotSS} performs \acrshort{fss} by projecting each pixel of a query image in a learned feature space and then labelling each of them with the same label of the closest prototype as shown in \autoref{fig:prototypical}.

In this family of solutions, it is central to design representative prototypes and use reliable distance metrics that allow correct label assignments. This pivotal consideration hinges on two fundamental design choices: the computation of prototypes and selecting a suitable distance metric.


A commonly embraced approach for prototype generation is the application of \acrfull{map} \cite{zhang2020sg}. \acrshort{map} distils informative representations from the feature volume projected by support images through chosen feature extractors, typically ResNet50, ResNet101 \cite{he2016deep}, or VGG16 \cite{vgg}. This process entails computing the Hadamard product—an element-wise multiplication—between the feature volume and the ground truth mask for each element in the support set. The averaged results of the Hadamard product yield a single feature vector embodying the class prototype. Practically, \acrshort{map} aggregates feature vectors from spatial locations in the feature volume where the ground truth mask delineates the subject. 
While widely used backbones include ResNet50, ResNet101, and VGG16, a consensus on the preferred backbone has yet to emerge. Ongoing innovations in backbone design, such as combining multiple backbones \cite{catalano2024more} or designing new ones, hold promising for generating more informative embeddings, thereby potentially enhancing overall segmentation performance.


The second critical design choice involves selecting an appropriate distance metric to measure the similarity between prototypes and query features. Snell et al. advocate for Bregman distance divergences, such as the squared Euclidean distance. However, empirical findings by Wang et al. \cite{wang2019panet} suggest that a cosine distance metric offers greater stability and superior performance compared to Euclidean distance, arguing that the bounded nature of cosine similarity makes it more suitable for optimization, leading to its preference in subsequent works, including \cite{wang2019panet, tian2020prior, liu2020part, zhang2020sg, zhang2019canet}.

However, it is crucial to acknowledge certain limitations associated with prototype-based models in \acrshort{fss}. In applications where subjects can be viewed as compositions of multiple sub-parts, attempting to find a single prototype representing all parts may lead to suboptimal segmentation performances. For instance, a dog can be decomposed into a head, body, and limbs, each with distinct features. Acknowledging this limitation, Yang et al. \cite{yang2020prototype} introduced a prototype mixture model, representing a class with more than one prototype to better capture the distinct macro feature of a subject class.
Taking this concept further, Zhang et al. \cite{zhang2019pyramid} and Wang et al. \cite{wang2020few} extended the prototype strategy by adopting a graph-based solution. They leveraged a fully connected graph, where each node represents feature vectors from support and query images, and connection weights represent the similarities between two feature vectors. These connection weights were then employed for the pixel-level classification task.

In the pursuit of further enhancing the prototype strategy, another method with a number of prototypes greater than the number of classes was proposed by Li et al. \cite{Li2021FewshotSS}. They recognized that a prototypical model might get confused by similar classes present in the background of an image. To address this issue, they introduced a pseudo-class for areas of the image that correspond to high activation but are not labelled. This approach allows the model to exploit better the limited support set by inferring a new class.

To improve the separation of the prototypes, Okazawa \cite{interclass_prototype_relation} introduces a loss component to disincentive the cosine similarity between prototypes. In this way, the author aims to reduce the risk of computing prototypes that are too similar due to the sparse and low coverage of the latent space typical of the few-shot scenario.
Conversely, Fan et. \cite{self_support_fss} update the prototypes by selecting the query features that self-match the prototypes with high confidence. Such features, while representing high-confidence parts of the subject in the query, allow for the inclusion of features less represented in the query.

Another departure from prototypical networks is proposed by Wu et al. \cite{wu2021learning}, recognizing that different classes may exhibit similar middle-level features despite having dissimilar appearances. To capitalize on this insight, they introduced a Meta-class Memory Module that aligns query and supports middle-level features to enhance the final segmentation mask prediction. The module learns meta-class embeddings, enabling effective utilization of similar features among different classes.

\begin{figure}[t]
    \centering
    \includesvg[width=\linewidth]{svg/new_prototypical_2.svg}
    
    \caption{Prototypical networks: a shared feature extractor gets a feature volume from both the support set and query images. The \acrfull{map} module takes the feature volume from the support set and masks its ground truth with the Hadamard product $\odot$ to compute the class prototype. The prediction mask $\hat{M}_q$ is calculated as a metric between the vector at each spatial location in the query feature volume with the class prototype.
    }
    
    \label{fig:prototypical}
    
\end{figure}

In summary, the design of prototypes and the choice of a suitable distance metric stand as the cornerstone of prototypical learning in the context of \acrshort{fss}. The utilization of \acrshort{map} and the selection of an appropriate distance metric play pivotal roles in achieving robust and accurate segmentation models. However, it is imperative to acknowledge certain limitations associated with prototype-based models in scenarios where subjects exhibit compositional complexity. In these scenarios, innovative solutions such as prototype mixture models or graph-based approaches are needed.
As we delve into the challenges of \acrshort{fss}, a crucial aspect comes into focus: the exploration of optimal backbones for feature extraction. The selection of a backbone, be it a well-established one like ResNet50, ResNet101, or VGG16, or the innovation of new backbones, has a pivotal role in defining the feature space and influencing the quality of the extracted embeddings. This choice becomes a determining factor shaping the overall performance of the downstream segmentation task. Hence, delving into the possibilities of diverse backbones becomes not only a worthwhile endeavour but a promising avenue to unlock improved segmentation performance.

    \subsection{Latent space optimization}
The effectiveness of \acrshort{fss} model designs can often be related to their representation of classes and thus in the corresponding feature space. This section presents various approaches for latent space class representation in \acrshort{fss}. Firstly, we discuss Generative Adversarial Networks (GANs) and how their regular latent space can be exploited. Secondly, we introduce studies that have achieved results similar to those of GANs but using simpler networks based on contrastive learning. Finally, we explore works that model the variability of subject classes using probabilistic frameworks such as Variational Autoencoders (VAEs).

        \subsubsection{Generative models}
            In certain applications of object segmentation, it is possible to have plenty of data available, but not enough label information; this scenario is particularly common in tasks like part segmentation. For instance, while it is possible to access to numerous images of objects like faces or cars through public datasets, creating detailed masks for their individual parts often requires considerable effort due to the absence of pre-existing parts labels. To fully exploit already available datasets and minimize the need for new labels, some works \cite{zhang2021datasetgan,tritrong2021repurposing,saha2022ganorcon} have resorted to the framework of \acrshort{fss} and developed methods based on generative models. The core idea is that good performing GANs such as StyleGAN \cite{karras2019style} or StlyGAN2 \cite{karras2020analyzing} have an internal representation of images that effectively synthesize both semantic and geometrical information and disentangle the latent factors of variation. This intuition has been supported by Karras et al. in \cite{karras2019style} where the authors show that interpolating latent codes of different images produces a synthetic image that is still qualitatively good. Given these intuitions, the internal representation of a \acrshort{GAN} generated image should be more informative than the feature extracted by other means and thus lead to better segmentation performance.

With this reasoning, Zhang et al. in \cite{zhang2021datasetgan} and Tritrong et al. in \cite{tritrong2021repurposing} exploit the regularity of StyleGAN  \cite{karras2019style} and StyleGAN2 \cite{karras2020analyzing} latent spaces to predict part segmentation masks. In their models, the segmentation is performed by a \acrshort{fss} module which predicts the masks from the \acrshort{GAN}'s internal representation of a generated image.
The practical implementation of this design involves a sequential process. Firstly, in the presence of abundant unlabeled target domain images, the GAN is trained on this data to optimize its latent space. Subsequently, the training set for the \acrshort{fss} module is constructed by sampling $k$ generated images along with their corresponding internal representations. The $k$ images are then manually annotated by a human annotator, serving as the ground truth for training the \acrshort{fss} module. 
Once the entire model is trained, Tritrong in \cite{tritrong2021repurposing} proposes to perform latent code optimization to make the GAN generate an image as close as possible to the test image and record its internal representation. The generated image and its internal representation are finally fed to the segmentation module to predict the parts segmentation mask. A depiction of the inference procedure for this model is proposed in \autoref{fig:GAN_latent_code_optm}.

As noted by the authors themselves, the latent code optimization step is computationally expensive. Additionally, the results of their experiments also suggest that the generated image might still be dissimilar to the test image, leading to poor performance. Tritrong et al. in \cite{tritrong2021repurposing} and Zhang et al. in \cite{zhang2021datasetgan} propose to overcome such limitations by using the trained model to generate a large synthetic dataset as shown in \autoref{fig:GAN_dataset}. The process involves sampling random new images and predicting the parts segmentation masks via the \acrshort{fss} module as before. The new synthetic dataset, composed of the generated images and inferred masks, can later be used to train any off-the-shelf segmentation model. Tritrong et al. in \cite{tritrong2021repurposing} validate this approach by showing that generating a new parts segmentation dataset avoiding the latent code optimization step reduces inference time and improves segmentation performance.

GANs can have complex training procedures, but once trained, they can be used as feature extractors as in \cite{tritrong2021repurposing, saha2022ganorcon}, or to produce new synthetic data to mitigate the data scarcity regime such as in \cite{zhang2021datasetgan}.
The wide range of strategies in this family of approaches branch makes it hard to define a general training procedure as it is possible for conditional networks and prototypical networks where episodic learning described in \autoref{episodic_training} is standard practice. Moreover, GAN-based FSS models are not restricted to the conventional values of 1 or 5 shots used in conditional and prototypical networks. For instance, Zang et al. in \cite{zhang2021datasetgan} generate 20 images from the GAN to train an MLP \acrshort{fss} module, similarly, Tritrong et al. in  \cite{tritrong2021repurposing} experiments with 1, 5 and 10 images to train the \acrshort{fss} module, while Saha et al. in \cite{saha2022ganorcon} uses 20 shots.

In conclusion, while GANs have shown promise for \acrshort{fss}, particularly in scenarios with abundant unlabeled data, their complex and computationally intensive training process raises questions about alternative approaches. The benefits of GANs' internal representation and regularity come with challenges, notably in the latent code optimization step. Section \ref{contrastive} delves into works exploring simpler architectures focusing on achieving comparable results in \acrshort{fss} without the complexities associated with training GANs.

\begin{figure}[t]
    \centering
    \includesvg[width=\linewidth]{svg/new_ReporposingGAN_Inference.svg}
    
    \caption{Exploitation of \acrshort{GAN} regular latent representation for \acrshort{fss}. A  \acrshort{GAN} is employed to generate an image that closely resembles the test image by latent code optimization. Once the generated image is sufficiently similar to the test image, its internal representation is extracted and fed into a \acrshort{fss} module to predict the final segmentation mask  $\hat{M}_q$.
    }
    \label{fig:GAN_latent_code_optm}
    
\end{figure}

\begin{figure}[t]
    \centering
    \includesvg[width=\linewidth]{svg/new_auto_shot.svg}
    
    \caption{A generative model is used to create a synthetic dataset with a limited annotation effort. First a \acrshort{GAN} is trained on the target dataset. A small number of synthetic images are labelled and in couple with their internal representations, are used to train a \acrshort{fss} module. During inference, an unlimited number of images and internal representations can be generated from the \acrshort{GAN} and fed into the \acrshort{fss} module to produce masks. The new synthetic dataset formed by the generated images and their masks can be used to train any off-the-shelf segmentation model.
    }
    \label{fig:GAN_dataset}
    
\end{figure}
        \subsubsection{Contrastive learning} \label{contrastive}
            In the realm of \acrshort{fss}, the effectiveness of \acrshort{GAN}s as powerful tools for generating regular internal representations has been acknowledged. However, Saha et al. in \cite{saha2022ganorcon} highlight how \acrshort{GAN}s can be challenging and time-consuming to train and their performance may decrease when using latent space optimization to get a test image latent code. To address these issues, Saha et al. introduced the use of contrastive learning for the feature extractor, demonstrating that this technique can make the backbone consistently produce latent representations that outperform \acrshort{GAN}-based models, without increasing its complexity. 

Contrastive learning, as described in \cite{NIPS2014_07563a3f,chen2021exploring,bachman2019learning}, involves training a model on a self-supervised proxy task to make its internal representations invariant to augmentations. This is achieved by constructing a dataset of image pairs, where one image is taken from the original dataset and the other is either a transformed version of the same image or a different one. The model is then trained to classify whether the two images represent the same subject or not. Once the model is trained it can be used as a feature extractor by feeding it with the test image and recording its internal representation. Saha et al. use the feature volume extracted in this manner to predict the part segmentation masks using a U-Net \cite{ronneberger2015u} inspired network. This model achieves slightly better results than the \acrshort{GAN} based model proposed by  Zhang et al. \cite{zhang2021datasetgan} while being easier to train.

In conclusion, contrastive learning emerges as a versatile tool for regularization in feature spaces of models employing feature extractors. This technique, applicable to various design strategies in \acrshort{fss}, holds the potential to enhance performance without escalating model complexity, prompting further exploration in the field.
        \subsubsection{Variational Auto Encoders}
            Within the domain of \acrshort{fss}, the prevalence of prototypical networks has been a cornerstone for representation in multiple works. However, the deterministic nature of their feature extractor presents limitations, especially in dealing with intra-class variations and noise in one-shot scenarios. Additionally, compressing an entire class into a single deterministic prototype vector can result in the loss of important structural information of the objects in the scene. Aligning with the pursuit of a more regular latent representation Wang et al. \cite{wang2021variational} propose an innovative approach based on Variational Autoencoders (VAEs) \cite{kingma2013auto,rezende2014stochastic}. Departing from traditional prototypical learning, this model introduces a shift toward a probabilistic framework, learning a probability distribution of prototypes for a given class. This allows for a more comprehensive encoding of object variations. In a later work \cite{sun2021attentional}, Sun et al. further enhanced the model by incorporating a variational attention mechanism to highlight the foreground of the test image, resulting in improved overall segmentation performance.
        \subsection{Foundation Models}

Recent developments in the \acrshort{ai} community have seen the emergence of Foundation Models, which are versatile models trained on extensive datasets and capable of adapting to diverse downstream tasks \cite{bommasani2021opportunities}. Notably, while prominent examples such as BERT \cite{kenton2019bert} and GPT \cite{gpt,GPT3} originate from the \acrfull{nlp} community, the realm of \acrfull{cv} also boasts its own foundation models like CLIP \cite{radford2021learning} and SAM \cite{kirillov2023segany}. 

CLIP can be framed as a bridge between \acrshort{nlp} and \acrshort{cv} by establishing a unified embedding space where images and text are represented. Through contrastive learning, CLIP aligns image and text embeddings, facilitating cross-modal understanding and enabling tasks such as classification, retrieval, and generation.
On the other hand, Kirillov et al. developed Segment Anything (SAM), a segmentation foundation model based on a novel promptable segmentation framework. Thanks to a data engine collecting 11M image-mask pairs, SAM is trained on vast and diverse datasets, allowing it to segment any objects in visual contexts. SAM accepts handcrafted prompts and returns the expected mask, supporting both sparse (points, boxes, text) and dense (masks) prompts. SAM stands out as a generalist vision foundation model for image segmentation, offering flexibility through a diverse range of input prompts. However, it lacks the ability to recognize the type of each segmented object.

The release of vision foundation models has sparked excitement in the \acrshort{cv} community, including in  the \acrshort{fss} research stream. Foundation models present a novel opportunity for \acrshort{fss}, as they can be leveraged to segment classes for which tailored datasets are not available. Notably, strategies for utilizing vision foundation models in \acrshort{fss} can be drastically different from the one already presented in this paper so far and require a separate discussion. In addition, in this category, we can identify three major types: prompt engineering, multi-modal approaches, and generalist models. This section delves deeper into these strategies, highlighting their significance in advancing the field of computer vision.

\subsubsection{Prompt engineering}

In the realm of image segmentation, foundation models like SAM offer remarkable capabilities but rely on additional guidance in the form of prompts tailored to specific object classes. SAM, designed to accept prompts in various forms such as point boxes, text, and masks, has spurred research efforts aimed at enhancing its specialization through prompt engineering.

One notable approach in this domain is introduced by Liu et al. with Matcher \cite{liu2023matcher}, a model designed for one-shot segmentation, leveraging SAM as its foundational model. Matcher begins by embedding both the support and query images  (in this context called reference and target image) using the DINOv2 feature extractor \cite{oquab2024dinov}. Given the correspondence between spatial locations in the feature volume and patches from the images, Matcher computes a similarity matrix between features to identify similar patches in the images. Points within patches achieving high similarity scores are selected as prompts for SAM, whose predictions are aggregated to generate the final segmentation output.

Similarly, Zhang et al. propose PerSAM \cite{zhang2023personalize}, a training-free personalization approach for SAM. PerSAM customizes SAM using one-shot data, comprising a reference image and a rough mask of the desired concept. Initially, PerSAM generates a location confidence map for the target object in the test image based on feature similarities. Subsequently, it selects two points as positive-negative location priors based on confidence scores, encoding them as prompt tokens for SAM's decoder to perform segmentation.

Both Matcher and PerSAM heavily rely on the feature volumes of reference and test images to generate appropriate prompts, emphasizing the importance of flexible and informative embeddings, drawing thus similarities with prototypical learning discussed in previous subsection. Additionally, SAM's flexibility in accommodating various prompt types suggests the potential for further exploration, particularly regarding the utilization of other available prompt types. In the subsequent subsection, we delve into works that leverage text prompts as part of a multimodal approach.

\subsubsection{Multimodal}

Multimodal approaches harness the synergy between different data modalities to enhance understanding and performance. One prominent application is the alignment of visual and textual embeddings, a concept dating back to proposals as far as 2014 \cite{Kiros2014UnifyingVE,faghri2018vse,7298932,7780382}. These techniques facilitate correlating subjects in images with their textual descriptions, laying the groundwork for text-promptable segmentation models. These models, not requiring any specific training for a novel class fall closely to  \acrshort{fss}, albeit requiring a text description instead of a segmentation example.


The introduction of CLIP reinvigorated research in this field. CLIP's training on a large and diverse datasets makes it highly flexible, enabling zero-shot classification of images. This flexibility has inspired many authors to propose one-shot segmentation models where the prompt is a textual description of the subject.
Numerous works in this field  \cite{lueddecke22_cvpr,10164075,9878961,zhou2022zegclip} have adopted an encoder-decoder architecture. Typically, the encoder leverages a pre-trained CLIP to extract aligned visual and text embeddings. The alignment of these embeddings, originating from inputs of different natures, empowers the decoder module to predict segmentation masks by focusing its design on the inner product between the text embeddings and the visual embeddings. While multimodal models utilizing text prompts alleviate the need for extensive segmentation datasets, it's worth noting that for certain application-specific domains, describing targets as texts may pose challenges. For example textual ambiguity could render this approach less than ideal in such cases.

In conclusion, while the concept of aligning textual and visual embeddings is not new, recent advancements, particularly with models like CLIP, have propelled this research area forward. One-shot segmentation models driven by textual prompts hold promise for various applications, albeit with considerations for domain-specific challenges and ambiguities in textual descriptions.
\subsubsection{Generalist models}
Generalist models, often built upon foundation models, exhibit remarkable adaptability across diverse applications. Leveraging techniques like In-Context Learning \cite{NEURIPS2022_960a172b,brown2020language,hao2022language}, primarily developed within \acrshort{nlp}, these models can swiftly accommodate new tasks with minimal examples or prompts. This adaptability extends to \acrshort{fsl} techniques, enabling their application in \acrshort{fss} tasks.
In-context learning, pioneered in \acrshort{nlp}, utilizes language sequences as a versatile interface, facilitating rapid adaptation to various language-centric tasks with minimal examples. However, while In-Context Learning has flourished in NLP, its exploration in computer vision remains nascent due to significant task disparities.
Unlike \acrshort{nlp} tasks, which revolve around discrete language tokens, \acrshort{cv} tasks encompass diverse output representations, posing challenges in defining task prompts. Addressing this disparity, Wang et al.\cite{Wang_2023_CVPR,seggpt} introduced an approach where images serve as the interface for visual perception. Their models, Painter \cite{Wang_2023_CVPR} and its successor SegGPT \cite{seggpt}, adopt the principles of \acrshort{nlp} models, utilizing images both as prompts and outputs. By framing dense prediction tasks as image inpainting, these models leverage three-component tensor inputs: an image paired with its label (akin to a support set in FSS literature), and a target image for prediction. 
Similarly, Meng et al. proposed SEGIC \cite{Meng2023SEGICUT}, a model combining foundation models and in-context learning. It employs a pre-trained foundation model for feature extraction, followed by a novel ``correspondence discovery" step. This step establishes semantic and geometric correspondences between the in-context and target feature maps, providing crucial guidance for segmentation. Subsequently, ``in-context instructions" are extracted based on the in-context samples and these correspondences. Finally, a lightweight decoder leverages this information to generate the segmentation mask for the target image. A key feature of these models and training procedures is their ability to leverage larger pre-training datasets. To this end, to guarantee SegGPT its flexibility, a key factor is its pre-training on a large combination of existing datasets, including ADE20K, MS-COCO \cite{MS-COCO}, PASCAL VOC \cite{pascal-voc-2012}, Cityscapes \cite{cordts2016cityscapes}, LIP \cite{LIP}, PACO \cite{PACO}, CHASE\_DB1 \cite{Chase},  DRIVE \cite{1282003}, HRF \cite{hrf}, STARE \cite{stare},  iSAID \cite{iSAID}, and loveDA \cite{loveda}.

While foundational and generalist models hold promise for vision tasks, including \acrshort{fss}, further investigation is imperative. Their superiority over task-specific models, particularly in diverse domains, necessitates comprehensive validation. Moreover, challenges such as large and diverse training sets and training procedures, as well as domain-specific performance variations and parameter tuning complexities, warrant attention. Exploring alternatives like backbones based on space-state models, such as MAMBA \cite{mamba} and VisionMAMBA\cite{vim}, could offer simplified architectures without compromising performance, paving the way for more efficient vision models.
            
        

    
    
    
\section{Models evaluation}

To train, test, and evaluate the variety of \acrshort{fss} models presented in previous sections, it is crucial to have publicly available standard datasets with densely annotated images that allow episodic training. Furthermore, a common evaluation metric is essential to rank works in the literature. In this section, we first present the metrics used to evaluate segmentation models, followed by a description of the publicly available benchmark datasets for \acrshort{fss}. In addition, we present a review of results compiled from the papers of some of the most significant models in the literature, showcasing their performance on the presented common standard benchmarks. Finally, in \autoref{qualitative_results} we present a selection of predictions obtained under the same conditions to qualitatively compare predictions of different models in multiple backbone and training dataset combinations.
    
    \subsection{Metrics}
        As common in segmentation tasks, the evaluation metrics commonly found in the \acrshort{fss} literature are based on the \acrfull{iou}. Two widely used metrics in \acrshort{fss} are the \acrfull{miou} or the \acrfull{fbiou}.
Recalling that for the task of segmenting a subject of class $c$ as pixel-wise classification 
the \acrshort{iou} is defined as:

\begin{equation} \label{eq:IoU_l}
IoU_c= \frac{TP_c}{TP_c+FP_c+FN_c}
\end{equation}
\noindent
where $TP_c$, $FP_c$ and $FN_c$ are the number of pixels that are true positive, false positive and false negative respectively.
The \acrshort{miou} is defined as the average of \acrshort{iou} for all classes:
\begin{equation} \label{eq:mIou}
mIoU = \frac{1}{C}\sum_{c=i}^{l=C}{IoU_c}
\end{equation}
\noindent
where C is the number of classes. 

Another common metric found in \acrshort{fss} is \acrfull{fbiou}. The \acrshort{fbiou} is a variation of the \acrshort{miou} as it ignores categories by only considering foreground and background (C = 2). However, this approach has been criticized for its limitations. In particular, Zhang et al. \cite{zhang2019canet} argue that ignoring categories can lead to biased results when there is a class imbalance in the dataset, such as in MS COCO or PASCAL VOC. The \acrshort{fbiou} may also not accurately reflect the performance of a model,  as failing to segment small objects will not necessarily be reflected in \acrshort{fbiou} scores. The reason is that if the background occupies a large portion of the scene, a small object will have little influence over the \acrshort{fbiou}. 

    \subsection{Public Datasets}

While introducing the problem of \acrlong{fss}, Shaban et al. in \cite{BMVC2017_167} introduced a new dataset to simulate segmentation episodes based on the data of the original PASCAL VOC 2012 \cite{pascal-voc-2012} dataset  and its extended annotations in SDS \cite{extended_labels_pascal_voc} naming it PASCAL-5\textsuperscript{i}.
Given the set $L$ of all 20 subject classes from the PASCAL VOC 2012 dataset Shaban et al. in \cite{BMVC2017_167} first sample 5 classes to create the label set $L_{test}$ and uses the remaining 15 to build $L_{train}$. Sampling different classes for $L_{test}$ and $L_{train}$ allow to create 4 folds from the original dataset as we can build $L_{test} = \{4i+1, ..., 4i+5\}, i \in [0,3] $. Once $L_{test}$ and $L_{train}$ are defined, Shaban et al. in \cite{BMVC2017_167} form the train dataset $D_{train}$ by picking all the $(image,mask)$ pairs from the PASCAL VOC train subset that contains subject from  $L_{train}$. Similarly, the test dataset $D_{test}$ contains all $(image,mask)$ pairs in the validation subset of PASCAL VOC containing subjects from  $L_{test}$. The two dataset $D_{train}$ and $D_{test}$ are then used to sample trainig and testing episodes as explained in \autoref{episodic_training}.

The generalization capacity of a model is a crucial property in \acrshort{fss}, therefore, the few classes of PASCAL-5\textsuperscript{i} can make it unideal as a benchmark. To this end,  Nguyen et al. in \cite{nguyen2019feature} introduced a more challenging dataset named COCO-20\textsuperscript{i} where episodes are created similarly to PASCAL-5\textsuperscript{i}, but now  with 80 classes based on MS COCO \cite{MS-COCO}. The 80 classes are sampled with the same procedure as for PASCAL-5\textsuperscript{i} leading to 4 possible folds of 20 class each from the original dataset.
While COCO-20\textsuperscript{i} has more classes than PASCAL-5\textsuperscript{i} Li et al. \cite{li2020fss} still found it inadequate to test \acrshort{fss} models as a benchmark dataset for this task should have a large number of classes while  requiring only few samples for class. To this end, they introduced a totally new dataset called FSS-1000 designed specifically for \acrshort{fss}, with 1000 classes having 10 images per class. Unlike  COCO-20\textsuperscript{i} and   PASCAL-5\textsuperscript{i}, this dataset only segments one class per image. This means that even if multiple class subjects are present in a single image, only one will be segmented, making it possible to evaluate only  \acrshort{fbiou} metric.

    \subsection{Benchmark results}  \label{quantitative_results}
        
        In this section, we present a comprehensive collection of benchmark results from various seminal works in the \acrshort{fss} field. By synthesizing these findings from the literature, we provide the reader with a consolidated resource, complete with details on the year and venue of each model's presentation. This presentation format facilitates easy comparison between different works and enables the observation of performance trends across recent years. Moreover, the tables presented herein offer insights into the attention received by different benchmark datasets, highlighting those that are more extensively explored and those that are less so. For instance, it becomes evident that while the older benchmark dataset PASCAL-5\textsuperscript{i} has been extensively tested, COCO-20\textsuperscript{i} has rapidly gained popularity, and FSS-1000 remains less prominent in the literature.

In reporting results for specialist \acrshort{fss} models, we provide metrics for both 1 Shot and 5 Shot settings, along with details on the backbone architecture utilized. Specifically, for benchmark datasets such as Pascal-5\textsuperscript{i} and COCO-20\textsuperscript{i}we report the \acrshort{miou} in \autoref{tab:pascal_results_miou} and \autoref{tab:coco_results_miou} and the \acrshort{fbiou} in \autoref{tab:pascal_results_fbiou} and \autoref{tab:coco_results_fbiou}, while for FSS-1000, we report the \acrshort{fbiou} in \autoref{tab:fss1000_results}. In table \autoref{table:pascal_specialit_generalist}, we report the scores of generalist models from the literature and compare them to some of the more modern specialist ones.

As is customary in the literature of  \acrshort{fss} for benchmark datasets where \acrshort{miou} is employed as a performance metric, we report the \acrshort{miou} on each fold as well as its average over the 4 folds. Conversely, folds are typically not used when evaluating performance with the \acrshort{fbiou} metric. As a result, we only report the performance metric on the entire dataset in such cases.

In essence, these tables serve not only as repositories of benchmark results but also enhance the value of the paper by providing a centralized and comparative analysis of performance across different models and datasets. This comprehensive analysis aids researchers in assessing the state-of-the-art and identifying promising avenues for future exploration.


    \begin{table}[H]
    \caption{Evaluation of 1-shot and  5-shot semantic segmentation specialist models on PASCAL-5\textsuperscript{i} using mean-IoU scores available in the literature. We report the average over the four folds, and, when available, the results on each individual fold. Results for PPNet without using additional unlabeled data are reported.}
    \resizebox{\textwidth}{!}{
        \begin{tabular}{ c|c|c|c|c|c|c|c|c|c|c|c } 
    
    \hline
    \multirow{2}{*}{Backbone}   & \multirow{2}{*}{Method} & \multicolumn{5}{|c|}{1 Shot} & \multicolumn{5}{|c}{5 Shot} \\
                                &                         & Fold-0 & Fold-1	& Fold-2 & Fold-3 & MIoU\% & Fold-0 & Fold-1	& Fold-2 & Fold-3 & MIoU\% \\
    
    \hline
    \multirow{16}{*}{VGG16}     
                                &  OSLSM \cite{BMVC2017_167}  (BMVC 2017) & 33.6 & 55.3 & 40.9 & 33.5 & 40.8 & 35.9 & 58.1 & 42.7 & 39.1 & 43.9 \\ 
                                &  PL + SEG \cite{Dong2018FewShotSS} (BMVC 2018) & - & - & - & - & 61.2 & - & - & - & - & 62.3 \\ 
                                &  co-FCN 2018 \cite{rakelly2018conditional} (ICLR 2018) & 36.7 & 50.6 & 44.9 & 32.4 & 41.1 & 37.5 & 50.0 & 44.1 & 33.9 & 41.4 \\ 
                                &  SG-One \cite{zhang2020sg} (TCYB 2019) & 40.2 & 58.4 & 48.4 & 38.4 & 46.3 & 41.9 & 58.6 & 48.6 & 39.4 & 47.1 \\ 
                                &  PANet \cite{wang2019panet} (ICCV 2019) & 42.3 & 58.0 & 51.1 & 41.2 & 48.1 & 51.8 & 64.6 & 59.8 & 46.5 & 55.7 \\ 
                                &  AMP \cite{siam2019adaptive} (ICCV 2019) & 41.9 & 50.2 & 46.7 & 34.7 & 43.4 & 41.8 & 55.5 & 50.3 & 39.9 & 46.9 \\ 
                                &  FWB \cite{nguyen2019feature} (ICCV 2019) & 47.0 & 59.6 & 52.6 & 48.3 & 51.9 & 50.9 & 62.9 & 56.5 & 50.1 & 55.1 \\ 
                                &  CRNet \cite{liu2020crnet} (CVPR 2020) & - & - & - & - & 55.2 & - & - & - & - & 58.5 \\ 
                                &  FSS-1000 \cite{li2020fss} (CVPR 2020) & - & - & - & - & - & 37.4 & 60.9 & 46.6 & 42.2 & 56.8 \\ 
                                &  PFENet \cite{tian2020prior} (TPAMI 2020) & 56.9 & 68.2 & 54.4 & 52.4 & 58.0 & 59.0 & 69.1 & 54.8 & 52.9 & 59.0 \\ 
                                &  RPMM \cite{liu2020part} (ECCV  2020) & 47.1 & 65.8 & 50.6 & 48.5 & 53.0 & 50.0 & 66.5 & 54.9 & 47.6 & 54.0 \\ 
                                &  SST \cite{zhu2020self} (IJCAI 2020) & 50.9 & 63.0 & 53.6 & 49.6 & 54.3 & 52.5 & 64.8 & 59.5 & 51.3 & 57.0 \\
                                &  HSNet \cite{min2021hypercorrelation} (ICCV 2021) & 59.6 & 65.7 & 59.6 & 54.0 & 59.7 & 64.9 & 69.0 & 64.1 & 58.6 & 64.1 \\ 
                                &  MM-Net \cite{wu2021learning} (ICCV 2021) & 57.1 & 67.2 & 56.6 & 52.3 & 58.3 & 56.6 & 66.7 & 53.6 & 56.5 & 58.3 \\ 
                                &  BAM \cite{lang2022learning} (CVPR 2022) & 63.2 & 70.8 & 66.1 & 57.5 & 64.4 & 67.4 & 73.1 & 70.6 & 64.0 & 68.8 \\ 
                                &  FECANet \cite{liu2023fecanet} (TMM 2023)   & 66.5 & 68.9 & 63.6 & 58.3 & 64.3 & 68.6 & 70.8 & 66.7 & 60.7 & 66.7 \\ 
    \hline
    \multirow{22}{*}{ResNet50}  
                                &    CANet \cite{zhang2019canet}  (ICCV 2019) & 52.5 & 65.9 & 51.3 & 51.9 & 55.4 & 55.5 & 67.8 & 51.9 & 53.2 & 57.1 \\ 
                                &    PGNet \cite{zhang2019pyramid} (ICCV 2019) & 56.0 & 66.9 & 50.6 & 50.4 & 56.0 & 57.7 & 68.7 & 52.9 & 54.6 & 58.5 \\ 
                                &    CRNet \cite{liu2020crnet} (CVPR 2020) & - & - & - & - & 55.7 & - & - & - & - & 58.8 \\ 
                                &    PMMs \cite{yang2020prototype} (ECCV 2020) & 52.0 & 67.5 & 51.5 & 49.8 & 55.2 & 55.0 & 68.2 & 52.9 & 51.1 & 56.8 \\ 
                                &    PPNet \cite{liu2020part}  (ECCV 2020) & 47.8 & 58.8 & 53.8 & 45.6 & 51.5 & 58.4 & 67.8 & 64.9 & 56.7 & 62.0 \\ 
                                &    RPMMs \cite{yang2020prototype} (ECCV 2020) & 55.2 & 66.9 & 52.6 & 50.7 & 56.3 & 56.3 & 67.3 & 54.5 & 51.0 & 57.3 \\ 
                                &    PFENet \cite{tian2020prior} (TPAMI 2020) & 61.7 & 69.5 & 55.4 & 56.3 & 60.8 & 63.1 & 70.7 & 55.8 & 57.9 & 61.9 \\ 
                                &    SST \cite{zhu2020self}  (IJCAI 2020) & 54.4 & 66.4 & 57.1 & 52.5 & 57.6 & 58.6 & 68.7 & 63.1 & 55.3 & 61.4 \\ 
                                &    SimPropNet \cite{SimPropNet} (IJCAI 2020) & 54.9 & 67.3 & 54.5 & 52.0 & 57.2 & 57.2 & 68.5 & 58.4  & 56.1 & 60.0 \\ 
                                &    LTM \cite{yang2020new} (MMM 2020) & 52.8 & 69.6 & 53.2 & 52.3 & 57.0 & 57.9 & 69.9 & 56.9 & 57.5 & 60.6 \\ 
                                &    ASGNet \cite{li2021adaptive} (CVPR 2021)& 58.8 & 67.9 & 56.8 & 53.7 & 59.3 & 63.7 & 70.6 & 64.2 & 57.4 & 63.9 \\ 
                                &    RePRI \cite{boudiaf2021few} (CVPR 2021) & 60.2 & 67.0 & 61.7 & 47.5 & 59.1 & 64.5 & 70.8 & 71.7 & 60.3 & 66.8 \\ 
                                &    CWT \cite{lu2021simpler} (ICCV 2021) & 56.3 & 62.0 & 59.9 & 47.2 & 56.4 & 61.3 & 68.5 & 68.5 & 56.6 & 63.7 \\ 
                                &    MM-Net \cite{wu2021learning} (ICCV 2021) & 62.7 & 70.2 & 57.3 & 57.0 & 61.8 & 62.2 & 71.5 & 57.5 & 62.4 & 63.4 \\ 
                                &    HSNet \cite{min2021hypercorrelation} (ICCV 2021) & 64.3 & 70.7 & 60.3 & 60.5 & 64.0 & 70.3 & 73.2 & 67.4 & 67.1 & 69.5 \\ 
                                &    VAT \cite{hong2021cost} (ECCV 2022) & 67.6 & 71.2 & 62.3 & 60.1 & 65.3 & 72.4 & 73.6 & 68.6 & 65.7 & 70.0 \\ 
                                &    BAM \cite{lang2022learning} (CVPR 2022) & 69.0 & 73.6 & 67.6 & 61.1 & 67.8 & 70.6 & 75.1 & 70.8 & 67.2 & 70.9 \\ 
                                &    DCAMA \cite{shi2022dense} (ECCV 2022) & 67.5 & 72.3 & 59.6 & 59.0 & 64.6 & 70.5 & 73.9 & 63.7 & 65.8 & 68.5 \\ 
                                &    IPMT \cite{NEURIPS2022_f7fef21d} (2022) & 72.8 & 73.7 & 59.2 & 61.6 & 66.8 & 73.1 & 74.7 & 61.6 & 63.4 & 68.2 \\ 
                                &    FECANet \cite{liu2023fecanet} (TMM 2023) & 69.2 & 72.3 & 62.4 & 65.7 & 67.4 & 72.9 & 74.0 & 65.2 & 67.8 & 70.0 \\ 
                                &    ABCNet \cite{Huang_2023_CVPR} (CVPR2023) & 68.8 & 73.4 & 62.3 & 59.5 & 66.0 & 71.7 & 74.2 & 65.4 & 67.0 & 69.6 \\ 
                                &    PMNet \cite{Chen_2024_WACV} (WACV 2024) & 67.3 & 72.0 & 62.4 & 59.9 & 65.4 & 73.6 & 74.6 & 69.9 & 67.2 & 71.3 \\ 
    \hline
    \multirow{16}{*}{ResNet101} 
                                &    FWB \cite{nguyen2019feature} (ICCV 2019) & 51.3 & 64.5 & 56.7 & 52.2 & 56.2 & 54.8 & 67.4 & 62.2 & 55.3 & 59.9 \\ 
                                &    PPNet \cite{liu2020part} (ECCV 2020) & 52.7 & 62.8 & 57.4 & 47.7 & 55.2 & 60.3 & 70.0 & 69.4 & 60.7 & 65.1 \\ 
                                &    DAN \cite{wang2020few} (ECCV 2020) & 54.7 & 68.6 & 57.8 & 51.6 & 58.2 & 57.9 & 69.0 & 60.1 & 54.9 & 60.5 \\ 
                                &    PFENet \cite{tian2020prior} (TPAMI 2020)  & 60.5 & 69.4 & 54.4 & 55.9 & 60.1 & 62.8 & 70.4 & 54.9 & 57.6 & 61.4 \\ 
                                &    VPI \cite{wang2021variational} (WACV 2021) & 53.4 & 65.6 & 57.3 & 52.9 & 57.3 & 55.8 & 67.5 & 62.6 & 55.7 & 60.4 \\ 
                                &    RePRI \cite{boudiaf2021few} (CVPR 2021) & 59.6 & 68.6 & 62.2 & 47.2 & 59.4 & 66.2 & 71.4 & 67.0 & 57.7 & 65.6 \\ 
                                &    HSNet \cite{min2021hypercorrelation} (ICCV 2021) & 67.3 & 72.3 & 62.0 & 63.1 & 66.2 & 71.8 & 74.4 & 67.0 & 68.3 & 70.4 \\ 
                                &    CWT \cite{lu2021simpler} (ICCV 2021) & 56.9 & 65.2 & 61.2 & 48.8 & 58.0 & 62.6 & 70.2 & 68.8 & 57.2 & 64.7 \\ 
                                &    CyCTR \cite{zhang2021few} (NIPs 2021) & 69.3 & 72.7 & 56.5 & 58.6 & 64.3 & 73.5 & 74.0 & 58.6 & 60.2 & 66.6 \\ 
                                &    ASGNet \cite{li2021adaptive}  (CVPR 2021) & 59.8 & 67.4 & 55.6 & 54.4 & 59.3 & 64.6 & 71.3 & 64.2 & 57.3 & 64.4 \\ 
                                &    VAT \cite{hong2021cost} (ECCV 2022) & 68.4 & 72.5 & 64.8 & 64.2 & 67.5 & 73.3 & 75.2 & 68.4 & 69.5 & 71.6 \\ 
                                &    DCAMA \cite{shi2022dense} (ECCV 2022) & 65.4 & 71.4 & 63.2 & 58.3 & 64.6 & 70.7 & 73.7 & 66.8 & 61.9 & 68.3 \\ 
                                &    IPMT \cite{NEURIPS2022_f7fef21d} (NeurIPS 2022) & 71.6 & 73.5 & 58.0 & 61.2 & 66.1 & 75.3 & 76.9 & 59.6 & 65.1 & 69.2 \\ 
                                &    API \cite{sun2021attentional} (Pattern Recognition 2023)  & 53.4 & 65.6 & 57.3 & 52.9 & 57.3 & 55.8 & 67.5 & 62.6 & 55.7 & 60.4 \\ 
                                &    ABCNet \cite{Huang_2023_CVPR} (CVPR 2023) & 65.3 & 72.9 & 65.0 & 59.3 & 65.6 & 71.4 & 75.0 & 68.2 & 63.1 & 69.4 \\ 
                                &    PMNet \cite{Chen_2024_WACV} (WACV 2024)           & 71.3 & 72.4 & 66.9 & 61.9 & 68.1 & 74.9 & 75.5 & 75.3 & 69.8 & 73.9 \\ 
    \hline
    \multirow{2}{*}{Swin-B} 
                                &    HSNet \cite{min2021hypercorrelation} (ICCV 2021)  & 67.9 & 74.0 & 60.3 & 67.0 & 67.3 & 72.2 & 77.5 & 64.0 & 72.6 & 71.6 \\
                                &    DCAMA \cite{shi2022dense} (ECCV 2022)             & 72.2 & 73.8 & 64.3 & 67.1 & 69.3 & 75.7 & 77.1 & 72.0 & 74.8 & 74.9 \\
    \hline

    \end{tabular}
    }
    
    \label{tab:pascal_results_miou}
\end{table}

\begin{table} [H]
    \caption{Evaluation of  1-shot and  5-shot semantic segmentation specialist models on PASCAL-5\textsuperscript{i} using FB-IoU scores available in the literature.}
    \centering
    \resizebox{!}{0.38\textheight}{
        \begin{tabular}{ c|c|c|c } 
    \hline
    Backbone   & Method & 1 Shot & 5 Shot \\        
    \hline
    \multirow{11}{*}{VGG16}   
                                & OSLSM \cite{BMVC2017_167} (BMVC 2017) & 61.3 & 61.5 \\
                                & co-FCN \cite{rakelly2018conditional} (ICLR 2018) & 60.1 & 60.2 \\
                                & PL + SEG \cite{Dong2018FewShotSS} (BMVC 2018) & 61.2 & 62.3 \\
                                & SG-One \cite{zhang2020sg} (TCYB 2019) & 63.9 & 65.9 \\
                                & PANet \cite{wang2019panet} (ICCV 2019) & 66.5 & 70.7 \\
                                & PFENet \cite{tian2020prior} (TPAMI 2020) & 72 & 72.3 \\
                                & HSNet \cite{min2021hypercorrelation} (ICCV 2021) & 73.4 & 76.6 \\
                                & BAM \cite{lang2022learning} (CVPR 2022) & 77.26 & 81.1 \\
                                & MMnet \cite{wu2021learning} (ICCV 2021) & 78.01 & 80.5 \\
                                & APANet \cite{chen2021apanet} (TMM 2022) & 71.3 & 75.2 \\
                                & AMP \cite{9009561} (ICCV 2019) & 62.2 & 63.8 \\
                                & FECANet \cite{liu2023fecanet} (TMM 2023) & 76.2  & 77.6 \\
    \hline
    \multirow{24}{*}{ResNet50}  
                                & CANet \cite{zhang2019canet} (ICCV 2019) & 66.2 & 69.6 \\
                                & PGNet \cite{zhang2019pyramid}  (ICCV 2019) & 69.9 & 70.5 \\
                                & PPNet \cite{liu2020part} (ECCV 2020) & 69.19 & 75.76 \\
                                & PFENet \cite{tian2020prior} (TPAMI 2020) & 73.3 & 73.9 \\
                                & HSNet \cite{min2021hypercorrelation} (ICCV 2021) & 76.7 & 80.6 \\
                                & BAM \cite{lang2022learning} (CVPR 2022) & 79.71 & 82.18 \\
                                & VAT \cite{hong2021cost} (ECCV 2022) & 77.4 & 80.9 \\
                                & MMnet \cite{wu2021learning} (ICCV 2021) & 80.44 & 83.23 \\
                                & SAGNN \cite{xie2021scale} (CVPR 2021) & 73.2 & 73.3 \\
                                & SCLNet \cite{Zhang_2021_CVPR} (CVPR 2021) & 71.9 & 72.8 \\
                                & APANet \cite{chen2021apanet} (TMM 2022) & 72.3 & 77.4 \\
                                & ASGNet \cite{li2021adaptive} (CVPR 2021) & 69.2 & 74.2 \\
                                & CMN \cite{Xie_2021_ICCV} (ICCV 2021) & 72.3 & 72.8 \\
                                & MANet \cite{ao2022few} (arXiv 2021) & 71.4 & 75.2 \\
                                & DCP \cite{ijcai2022p143} (IJCAI 2022) & 75.6 & 79.7 \\
                                & MFGN \cite{electronics11193195} (MDPI 2022) & 76.4 & 79.3 \\
                                & CRNet \cite{liu2020crnet} (CVPR 2020) & 66.8 & 71.5 \\
                                & CMNet \cite{9813426} (TMM 2022) & 73.5 & 74.1 \\
                                & SimPropNet \cite{SimPropNet} (IJCAI 2020) & 73 & 72.9 \\
                                & CMN \cite{9711406} (ICCV 2021) & 72.3 & 72.5 \\
                                & ASNet \cite{kang2022integrative} (CVPR 2022) & 77.7 & 80.4 \\
                                & SCL(CANet) \cite{zhangself} (CVPR 2021) & 70.3 & 70.7 \\
                                & SCL(PFENet) \cite{zhangself} (CVPR 2021) & 71.9 & 72.8 \\
                                & IPMT \cite{NEURIPS2022_f7fef21d} (NeurIPS 2022) & 77.1 & 81.4\\
                                & FECANet \cite{liu2023fecanet} (TMM 2023) & 78.7  & 80.7 \\
    \hline
    \multirow{11}{*}{ResNet101} 
                                & A-MCG \cite{hu2019attention} (AAAI 2019) & 61.2 & 62.2 \\
                                & PFENet \cite{tian2020prior} (TPAMI 2020) & 72.9 & 73.5 \\
                                & PPNet \cite{liu2020part} (ECCV 2020) & 70.9 & 77.5 \\
                                & DAN \cite{wang2020few} (ECCV 2020) & 71.9 & 72.3 \\
                                & HSNet \cite{min2021hypercorrelation} (ICCV 2021) & 77.6 & 80.6 \\
                                & CyCTR \cite{zhang2021few} (NIPs 2021) & 72.9 & 75 \\
                                & VAT \cite{hong2021cost} (ECCV 2022) & 78.8 & 82 \\
                                & APANet \cite{chen2021apanet} (TMM 2022) & 74.9 & 78.8 \\
                                & MFGN \cite{electronics11193195} (MDPI 2022) & 76.8 & 79.6 \\
                                & ASGNet \cite{li2021adaptive} (CVPR 2021) & 71.7 & 75.2 \\
                                & IPMT \cite{NEURIPS2022_f7fef21d} (NeurIPS 2022) & 78.5 & 80.3\\
        \hline
    \multirow{2}{*}{Swin-B} 
                                & HSNet \cite{min2021hypercorrelation} (ICCV 2021)  & 77.9 & 81.2 \\
                                & DCAMA \cite{shi2022dense} (ECCV 2022)             & 78.5 & 82.9 \\
    \hline
    \end{tabular}
    }
    
    \label{tab:pascal_results_fbiou}
\end{table}

\begin{table}[H]
    \caption{Evaluation of  1-shot and   5-shot semantic segmentation specialist models on MS COCO-20\textsuperscript{i} using mean-IoU scores available in the literature. We report the average over the four folds and, when available, the results for each individual fold. Results for PPNet without using additional unlabeled data are reported.}
     \resizebox{\textwidth}{!}{
            \begin{tabular}{ c|c|c|c|c|c|c|c|c|c|c|c } 
            \hline
        
        \multirow{2}{*}{Backbone}   & \multirow{2}{*}{Method} & \multicolumn{5}{|c|}{1 Shot} & \multicolumn{5}{|c}{5 Shot} \\
                                    &                         & Fold-0 & Fold-1	& Fold-2 & Fold-3 & MIoU\% & Fold-0 & Fold-1	& Fold-2 & Fold-3 & MIoU\% \\
        
        \hline
        \multirow{4}{*}{VGG16} 
    
            & PANet2019 \cite{wang2019panet} (ICCV 2019)        & 28.7 & 21.2 & 19.1 & 14.8 & 20.9 & 39.4 & 28.3 & 28.2 & 22.7 & 29.7 \\
            & FWB \cite{nguyen2019feature} (ICCV 2019)          & 18.4 & 16.7 & 19.6 & 25.4 & 20.0 & 20.9 & 19.2 & 21.9 & 28.4 & 22.6 \\
            & PFENet \cite{tian2020prior} (TPAMI 2020)          & 35.4 & 38.1 & 36.8 & 34.7 & 36.3 & 38.2 & 42.5 & 41.8 & 38.9 & 40.4 \\
            & BAM \cite{lang2022learning} (CVPR 2022)           & 39.0 & 47.0 & 46.4 & 41.6 & 43.5 & 47.0 & 52.6 & 48.6 & 49.1 & 49.3 \\
    
        \hline
        \multirow{21}{*}{ResNet50}
            & PANet \cite{wang2019panet} (ICCV 2019)            & 31.5 & 22.6 & 21.5 & 16.2 & 23.0 & 45.9 & 29.2 & 30.6 & 29.6 & 33.8 \\
            & PMMs \cite{yang2020prototype} (ECCV 2020)         & 29.3 & 34.8 & 27.1 & 27.3 & 29.6 & 33.0 & 40.6 & 30.3 & 33.3 & 34.3 \\
            & BriNet \cite{yang2020brinet} (BMVC2020)           & 32.9 & 36.2 & 37.4 & 30.9 & 34.4 & -    & -    & -    & -    & -    \\
            & RPMMs \cite{yang2020prototype} (ECCV 2020)        & 29.5 & 36.8 & 29.0 & 27.0 & 30.6 & 33.8 & 42.0 & 33.0 & 33.3 & 35.5 \\
            & PPNet \cite{liu2020part}  (ECCV 2020)             & 34.5 & 25.4 & 24.3 & 18.6 & 25.7 & 48.3 & 30.9 & 35.7 & 0.3  & 36.2 \\
            & HFA \cite{9356450}  (TIP 2021)                    & 28.7 & 36.0 & 30.2 & 33.3 & 32.0 & 32.7 & 42.1 & 30.4 & 36.2 & 35.3 \\
            & MM-Net \cite{wu2021learning} (ICCV 2021)          & 34.9 & 41.0 & 37.2 & 37.0 & 37.5 & 37.0 & 40.3 & 39.3 & 36.0 & 38.2 \\
            & RePRI \cite{boudiaf2021few} (CVPR 2021)           & 31.2 & 38.1 & 33.3 & 33.0 & 34.0 & 38.5 & 46.2 & 40.0 & 43.6 & 42.1 \\
            & HSNet \cite{min2021hypercorrelation} (ICCV 2021)  & 36.3 & 43.1 & 38.7 & 38.7 & 39.2 & 43.3 & 51.3 & 48.2 & 45.0 & 46.9 \\
            & CyCTR \cite{zhang2021few} (NIPs 2021)             & 38.9 & 43.0 & 39.6 & 39.8 & 40.3 & 41.1 & 48.9 & 45.2 & 47.0 & 45.6 \\
            & ASGNet \cite{li2021adaptive} (CVPR 2021)          & -    & -    & -    & -    & 34.6 & -    & -    & -    & -    & 42.5 \\
            & CMN  \cite{xie2021few} (ICCV 2021)                & 37.9 & 44.8 & 38.7 & 35.6 & 39.3 & 42.0 & 50.5 & 41.0 & 38.9 & 43.1 \\
            & CWT \cite{lu2021simpler} (ICCV 2021)              & 32.2 & 36.0 & 31.6 & 31.6 & 32.9 & 40.1 & 43.8 & 39.0 & 42.4 & 41.3 \\
            & MFGN \cite{electronics11193195} (MDPI 2022)       & 40.8 & 45.5 & 41.1 & 39.1 & 41.6 & 46.1 & 52.3 & 46.2 & 44.3 & 47.2 \\
            & VAT \cite{hong2021cost} (ECCV 2022)               & 39.0 & 43.8 & 42.6 & 39.7 & 41.3 & 44.1 & 51.1 & 50.2 & 46.1 & 47.9 \\
            & BAM \cite{lang2022learning} (CVPR 2022)           & 43.4 & 50.6 & 47.5 & 43.4 & 46.2 & 49.3 & 54.2 & 51.6 & 49.6 & 51.2 \\
            & IPMT \cite{NEURIPS2022_f7fef21d} (NeurIPS 2022)   & 41.4 & 45.1 & 45.6 & 40.0 & 43.0 & 43.5 & 49.7 & 48.7 & 47.9 & 47.5 \\ 
            & DCAMA \cite{shi2022dense} (ECCV 2022)             & 41.9 & 45.1 & 44.4 & 41.7 & 43.3 & 45.9 & 50.5 & 50.7 & 46.0 & 48.3 \\ 
            & FECANet \cite{liu2023fecanet} (TMM 2022)          & 38.5 & 44.6 & 42.6 & 40.7 & 41.6 & 44.6 & 51.5 & 48.4 & 45.8 & 47.6 \\ 
            & ABCNet \cite{Huang_2023_CVPR} (CVPR 2023)         & 42.3 & 46.2 & 46.0 & 42.0 & 44.1 & 45.5 & 51.7 & 52.6 & 46.4 & 49.1 \\ 
            & PMNet \cite{Chen_2024_WACV} (WACV 2024)           & 39.8 & 41.0 & 40.1 & 40.7 & 40.4 & 50.1 & 50.0 & 49.6 & 49.6 & 50.3 \\ 

                  \hline
        \multirow{10}{*}{ResNet101}
            & FWB \cite{nguyen2019feature} (ICCV 2019)          & 17.0 & 18.0 & 21.0 & 28.9 & 21.2 & 19.1 & 21.5 & 23.9 & 30.1 & 23.7 \\
            & DAN \cite{wang2020few} (ECCV 2020)                & -    & -    & -    & -    & 24.4 & -    & -    & -    & -    & 29.6 \\
            & PFENet \cite{tian2020prior} (TPAMI 2020)          & 36.8 & 41.8 & 38.7 & 36.7 & 38.5 & 40.4 & 46.8 & 43.2 & 40.5 & 42.7 \\
            & HSNet \cite{min2021hypercorrelation} (ICCV 2021)  & 37.2 & 44.1 & 42.4 & 41.3 & 41.2 & 45.9 & 53.0 & 51.8 & 47.1 & 49.5 \\
            & VPI \cite{wang2021variational} (WACV 2021)        & 24.9 & 25.8 & 21.9 & 21.1 & 23.4 & 27.8 & 29.3 & 24.7 & 29.4 & 27.8 \\
            & CWT \cite{lu2021simpler} (ICCV 2021)              & 30.3 & 36.6 & 30.5 & 32.2 & 32.4 & 38.5 & 46.7 & 39.4 & 43.2 & 42.0 \\
            & API \cite{sun2021attentional} (Pattern Recognition 2023)        & 33.9     & 46.6     & 24.8     & 31.2     & 31.2     & 35.7 & 47.5 & 31.6 & 35.9 & 35.9 \\
            & IPMT \cite{NEURIPS2022_f7fef21d} (NeurIPS 2022)   & 40.5 & 45.7 & 44.8 & 39.3 & 42.6 & 45.1 & 50.3 & 49.3 & 46.8 & 47.9 \\ 
            & DCAMA \cite{shi2022dense} (ECCV 2022)             & 41.5 & 46.2 & 45.2 & 41.3 & 43.5 & 48.0 & 58.0 & 54.3 & 47.1 & 51.9 \\ 
            & PMNet \cite{Chen_2024_WACV} (WACV 2024)           & 44.7 & 44.3 & 44.0 & 41.8 & 43.7 & 52.6 & 53.3 & 53.5 & 52.8 & 53.1 \\ 
              \hline

        \multirow{2}{*}{Swin-B} 
            & HSNet \cite{min2021hypercorrelation} (ICCV 2021)  & 43.6 & 49.9 & 49.4 & 46.4 & 47.3 & 50.1 & 58.6 & 56.7 & 55.1 & 55.1 \\
            & DCAMA \cite{shi2022dense} (ECCV 2022)             & 49.5 & 52.7 & 52.8 & 48.7 & 50.9 & 55.4 & 60.3 & 59.9 & 57.5 & 58.3 \\
        \hline
    \end{tabular}
    }
    \label{tab:coco_results_miou}
    \end{table}

    \begin{table}[H]
        \caption{Evaluation of  1-shot and  5-shot semantic segmentation specialist models on COCO-20\textsuperscript{i} using FB-IoU scores available in the literature.}
        \centering
        \resizebox{!}{!}{
            \begin{tabular}{ c|c|c|c } 
        \hline
        Backbone   & Method & 1 Shot & 5 Shot \\        
        \hline
        \multirow{4}{*}{VGG16}       
                                    & PANet \cite{wang2019panet} (ICCV 2019) & 59.2 & 63.5 \\
                                    & PFENet \cite{tian2020prior} (TPAMI 2020) & 60 & 61.6 \\
                                    & SAGNN \cite{xie2021scale} (CVPR 2021) & 61.2 & 63.1 \\
                                    & FECANet \cite{liu2023fecanet} (TMM 2022) & 65.5 & 67.7 \\
        \hline
        \multirow{10}{*}{ResNet50}  
                                    & ASGNet \cite{li2021adaptive} (CVPR 2021) & 60.4 & 66.9 \\
                                    & HSNet \cite{min2021hypercorrelation} (ICCV 2021) & 68.2 & 70.7 \\
                                    & VAT \cite{hong2021cost} (ECCV 2022) & 68.8 & 72.4 \\
                                    & CMN \cite{9711406} (ICCV 2021) & 61.7 & 63.3 \\
                                    & MFGN \cite{electronics11193195} (MDPI 2022) & 65.2 & 69.1 \\
                                    & CMN \cite{9711406} (ICCV 2021) & 61.7 & 63.3 \\
                                    & ASNet \cite{kang2022integrative} (CVPR 2022) & 68.8 & 71.6 \\
                                    & FECANet \cite{liu2023fecanet} (TMM 2022) & 69.6 & 71.1 \\
        \hline
        \multirow{6}{*}{ResNet101} 
                                    & DAN \cite{wang2020few} (ECCV 2020) & 62.3 & 63.9 \\
                                    & PFENet \cite{tian2020prior} (TPAMI 2020) & 63 & 65.8 \\
                                    & HSNet \cite{min2021hypercorrelation} (ICCV 2021) & 69.1 & 72.4 \\
                                    & MFGN \cite{electronics11193195} (MDPI 2022) & 65.6 & 69.1 \\
                                    & SAGNN \cite{xie2021scale} (CVPR 2021) & 60.9 & 63.4 \\
                                    & A-MCG \cite{hu2019attention} (AAAI 2019) & 52 & 54.7 \\
            \hline
        \multirow{2}{*}{Swin-B} 
                                    & HSNet \cite{min2021hypercorrelation} (ICCV 2021)  & 72.5 & 76.1 \\
                                    & DCAMA \cite{shi2022dense} (ECCV 2022)             & 73.2 & 76.9 \\
        \hline
        \end{tabular}
        }
        
        \label{tab:coco_results_fbiou}
    \end{table}

\begin{table} [H]
    \caption{Quantitative comparison results of semantic segmentation specialist models on FSS-1000 dataset with the \acrshort{miou} scores available in the literature.}
    \centering
    \resizebox{!}{!}{
        \begin{tabular}{ c|c|c|c } 
    \hline
    Backbone   & Method & 1 Shot & 5 Shot \\        
    \hline
    \multirow{7}{*}{VGG16}   
                                & OSLSM \cite{BMVC2017_167} (BMVC 2017) & 70.3 & 73 \\
                                & co-FCN \cite{rakelly2018conditional} (ICLR 2018) & 71.2 & 74.2 \\
                                & FSS-1000 \cite{li2020fss} (CVPR 2020) & 73.4 & 80.1 \\ 
                                & DoG-LSTM \cite{azad2021texture} (WACV 2021)  & 80.8 & 83.4 \\
                                & HSNet \cite{min2021hypercorrelation} (ICCV 2021) & 82.3 & 85.8 \\
                                & API \cite{sun2021attentional} (arXiv 2021) & 83.4 & 85.3 \\
                                & DCAMA \cite{shi2022dense} (ECCV 2022) & 88.2 & 88.8 \\
    \hline
    \multirow{7}{*}{ResNet50} 
                                & FOMAML \cite{hendryx2019meta} (arXiv 2019) & 75.1 & 80.6 \\
                                & FSOT \cite{9813426} (TMM 2022) & 82.5 & 83.8 \\
                                & HSNet \cite{min2021hypercorrelation} (ICCV 2021) & 85.5 & 87.8 \\
                                & VAT \cite{hong2021cost} (ECCV 2022) & 89.5 & 90.3 \\
                                & ASNet \cite{kang2022integrative} (CVPR 2022) & 68.8 & 71.6 \\
                                & DCAMA \cite{shi2022dense} (ECCV 2022) & 92.4 & 93.1\\
                                & PMNet \cite{Chen_2024_WACV} (WACV 2024) & 84.6 & 86.3\\
    \hline
    \multirow{5}{*}{ResNet101} 
                                & DAN \cite{wang2020few} (ECCV 2020) & 85.2 & 88.1 \\
                                & HSNet \cite{min2021hypercorrelation} (ICCV 2021) & 86.5 & 88.5 \\
                                & VAT \cite{hong2021cost} (ECCV 2022) & 90 & 90.6 \\
                                & API \cite{sun2021attentional} (arXiv 2021) & 85.6 & 88 \\
                                & DCAMA \cite{shi2022dense} (ECCV 2022) & 88.3 & 89.1\\
    \hline
    \multirow{2}{*}{Swin-B} 
                                & HSNet \cite{min2021hypercorrelation} (ICCV 2021)  & 86.7 & 88.9\\
                                & DCAMA \cite{shi2022dense} (ECCV 2022)             & 90.1 & 90.4\\
    \hline
    \end{tabular}
    }
    \label{tab:fss1000_results}
\end{table}

\begin{table} [H]
    \centering
    \caption{Quantitative comparison between some modern specialist and generalist models on PASCAL-5$^i$, COCO-20$^i$, and FSS-1000. The results report the mean-IoU scores available in the literature.}
    \small 
    \begin{tabular}{c|c|c|c|c|c|c}
    \hline
    \multirow{2}{*}{Method}                                             & \multicolumn{2}{c|}{PASCAL-5$^i$}      & \multicolumn{2}{c|}{COCO-20$^i$}     & \multicolumn{2}{c}{FSS-1000}     \\
                                                                        & one-shot     & few-shot                & one-shot     & few-shot             & one-shot     & few-shot  \\
    \hline
                    \textit{\small specialist model}&&&&&&\\
                    HSNet \cite{min2021hypercorrelation} (ICCV 2021)    & 67.3         & 71.6                    & 47.3         & 55.1                 & 86.7        & 88.9 \\
                    DCAMA \cite{shi2022dense} (ECCV 2022)               & 69.3         & 74.9                    & 50.9         & 58.3                 & 90.1        & 90.4 \\
                    VAT \cite{hong2021cost} (ECCV 2022)                 & 67.9         & 72.0                    & 41.3         & 47.9                 & 90.0        & 90.6 \\
                    PMNet \cite{Chen_2024_WACV} (WACV 2024)             & 68.1         & 73.9                    & 43.7         & 53.1                 & 84.6        & 86.3 \\
    \hline
    \textit{\small generalist model} &&&&&&\\
                    Painter \cite{Wang_2023_CVPR} (CVPR 2023)           & 64.5         & 64.6                    & 32.8         & 32.6                 & 61.7        & 62.3\\
                    SegGPT  \cite{seggpt} (ICCV 2023)                       & 83.2         & 89.8                    & 56.1         & 67.9                 & 85.6        & 89.3\\
    \hline
    \end{tabular}

    \label{table:pascal_specialit_generalist}
\end{table}


    \subsection{Qualitative samples} \label{qualitative_results}

In the \acrshort{fss} field, researchers often present their models with various combinations of backbones and training sets, leading to diverse strengths and weaknesses for each combination. While quantitative results, as presented in the tables of Section~\ref{quantitative_results}, offer valuable insights into model performance, they may not fully convey the nuanced capabilities of different model configurations. To provide a more intuitive understanding of how these combinations influence a model's effectiveness and to facilitate qualitative comparisons between different models, we present \autoref{tab_qualitative_samples}. This table showcases the one-shot predictions of selected models under various configurations while reporting the achieved \%IoU score achieved by each prediction, contextualizing the quantitative results reported in Section~\ref{quantitative_results}.

From this qualitative analysis, notable observations emerge. For instance, it becomes evident that certain classes pose greater challenges for prediction, while some models demonstrate higher resilience to disparities between support and query images. A closer examination of specific rows in the table reveals insightful patterns. Notably, in the second row, all models struggle to differentiate between cars and buses. Furthermore, disparities in appearance from the support to the query image, such as those seen in the chair and person in rows 3 and 8, disproportionately affect the performance of certain models.
Here is also evident how the generalist model SegGPT effectively leverages its larger and composite pretraining dataset, in many cases predicting higher quality masks with respect to its specialist counterparts, casting light on a possible direction for this research field. 

In conclusion, qualitative analysis of predictions is indispensable for assessing the reliability of a model in specific applications or under particular challenges. Such analysis can unveil insights that may be challenging to glean solely from quantitative results.


    \begin{table}[H]
        \centering
        \caption{Qualitative 1-shot prediction samples from a selection of models. The first two columns show the support image with the support mask overlayed in red; the second column shows the query image along with the ground truth mask in overlay; the rest of the table shows the prediction by the selected models. For each prediction of the specialist models, the top row indicates the name of the model, the second reports the name of the backbone, and the third one reports the dataset used for training. For the generalist model SegGPT is not possible to talk about a backbone, and its training set is composed of a large combination of datasets. Different columns for a single model allow comparing predictions when using different combinations of backbones and training datasets. Under each prediction is reported the \%IoU with respect of the ground truth.
}
        \def\arraystretch{0.5}
        \renewcommand{\arraystretch}{1.5} 
        \setlength\tabcolsep{1pt}
        \begin{tabular}{cc|cc|cc|cc|cc|cc|c|c}
\multirow{3}{*}{Support}& \multirow{3}{*}{Query}& \multicolumn{6}{c|}{DCAMA}                                                                    & \multicolumn{2}{c|}{HSNe}       & \multicolumn{3}{c|}{VAT}                       & SegGPT  \\
                        &                       & \multicolumn{2}{c|}{ResNet101} & \multicolumn{2}{c|}{ResNet50}  & \multicolumn{2}{c|}{Swin-B} & \multicolumn{2}{c|}{ResNet101}  & \multicolumn{2}{c|}{ResNet101}   &  ResNet50  &         \\
                        &                       & \tiny COCO & \tiny PASCAL      & \tiny COCO     & \tiny PASCAL  & \tiny COCO  & \tiny PASCAL  & \tiny COCO      & \tiny PASCAL  & \tiny FSS-1000    & \tiny PASCAL & \tiny COCO &      \\
                        \includegraphics[width=0.065\linewidth,  valign=m]{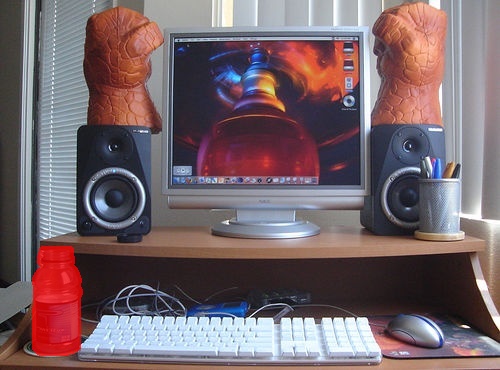} &
                        \includegraphics[width=0.065\linewidth,  valign=m]{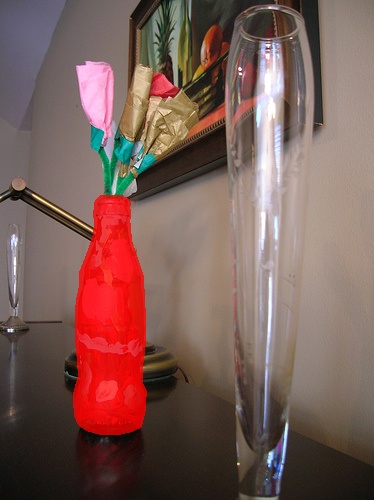} &
                        \includegraphics[width=0.065\linewidth,  valign=m]{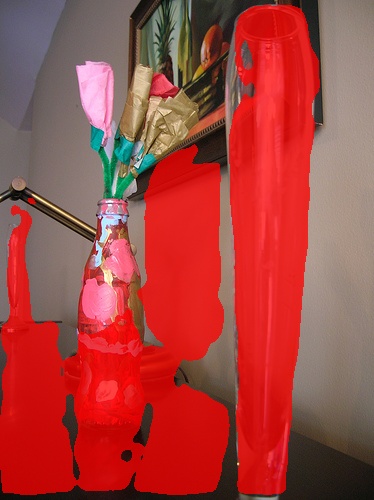} &
                        \includegraphics[width=0.065\linewidth,  valign=m]{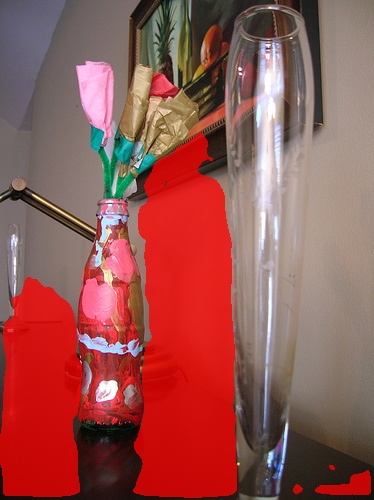} &
\includegraphics[width=0.065\linewidth,  valign=m]{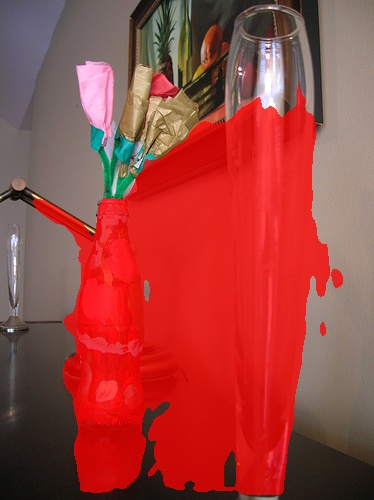} &
\includegraphics[width=0.065\linewidth,  valign=m]{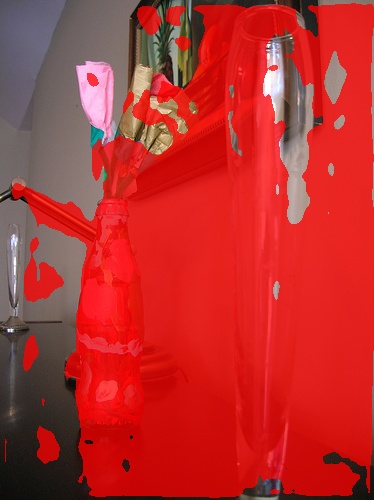} &
\includegraphics[width=0.065\linewidth,  valign=m]{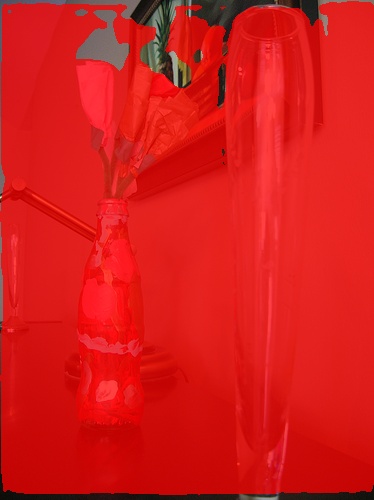} &
\includegraphics[width=0.065\linewidth,  valign=m]{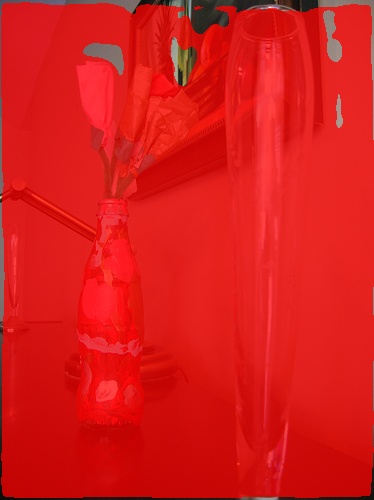} &
\includegraphics[width=0.065\linewidth,  valign=m]{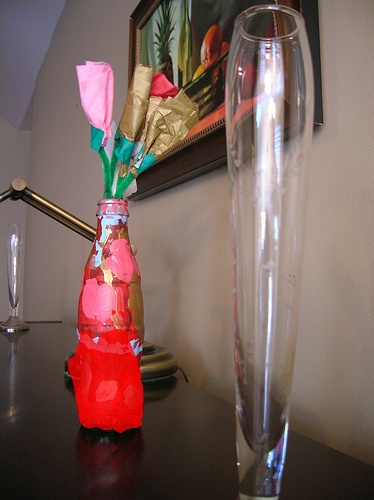} &
\includegraphics[width=0.065\linewidth,  valign=m]{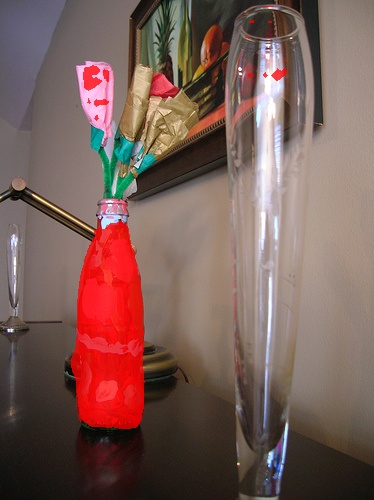} &
\includegraphics[width=0.065\linewidth,  valign=m]{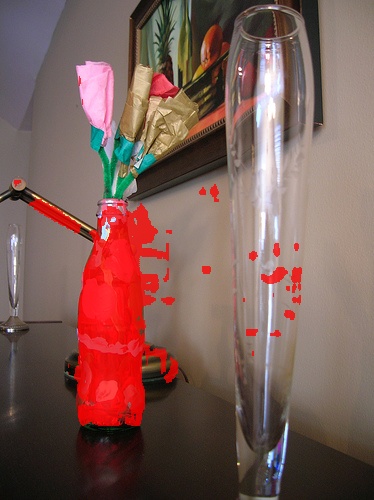} &
\includegraphics[width=0.065\linewidth,  valign=m]{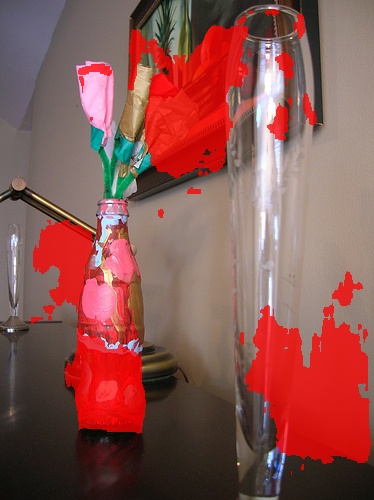} &
\includegraphics[width=0.065\linewidth,  valign=m]{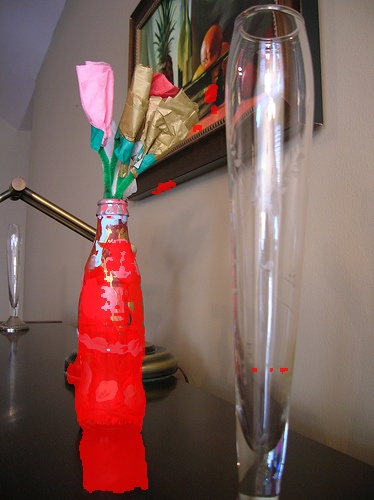} &
\includegraphics[width=0.065\linewidth,  valign=m]{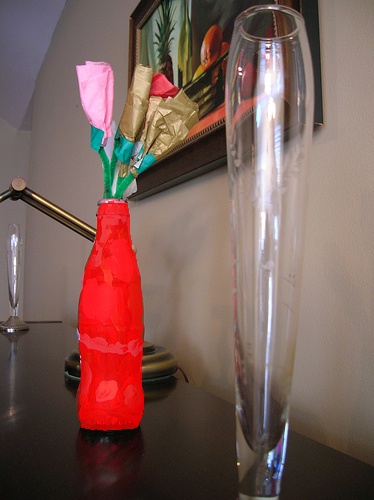} \\
& & 8.25 & 0.76 & 17.73 & 11.10 & 8.14 & 8.00 & 40.73 & 79.90 & 65.83 & 13.34 & 54.31 & 85.50		\\
\includegraphics[width=0.065\linewidth,  valign=m]{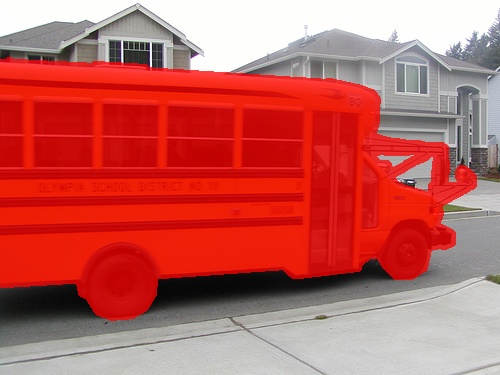} &
\includegraphics[width=0.065\linewidth,  valign=m]{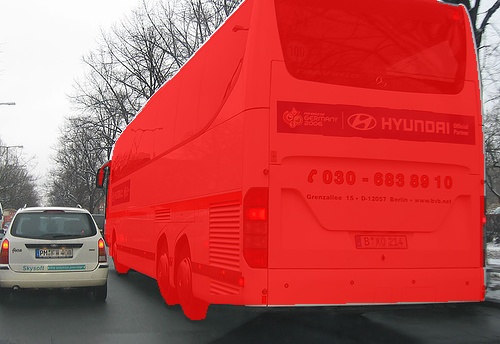} &
\includegraphics[width=0.065\linewidth,  valign=m]{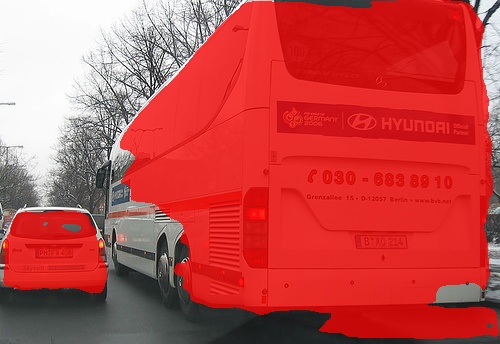} &
\includegraphics[width=0.065\linewidth,  valign=m]{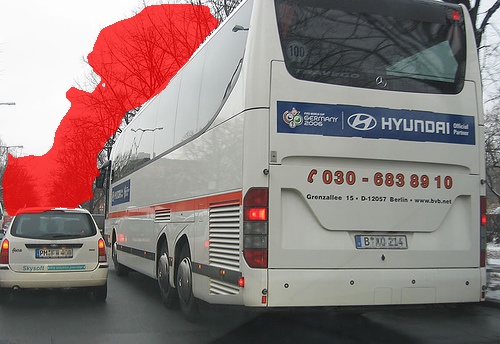} &
\includegraphics[width=0.065\linewidth,  valign=m]{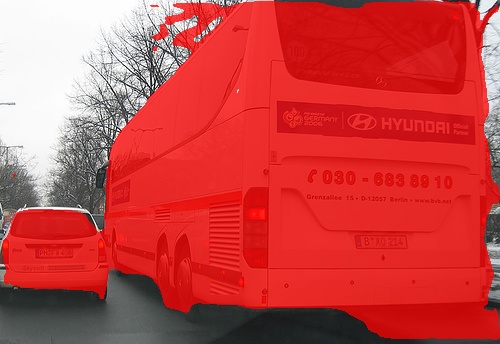} &
\includegraphics[width=0.065\linewidth,  valign=m]{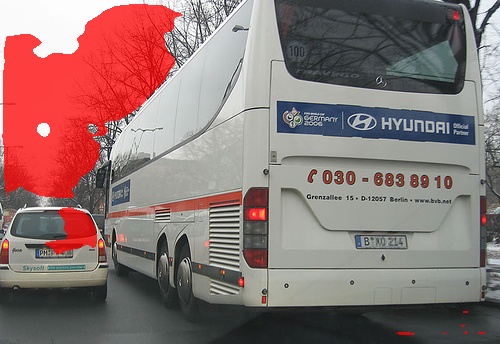} &
\includegraphics[width=0.065\linewidth,  valign=m]{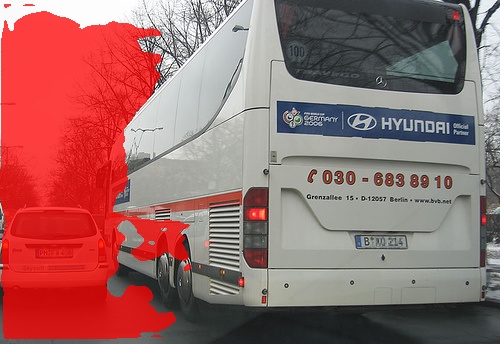} &
\includegraphics[width=0.065\linewidth,  valign=m]{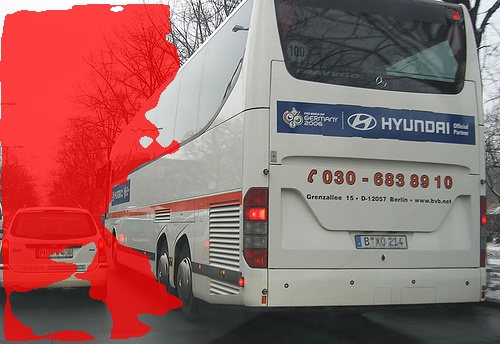} &
\includegraphics[width=0.065\linewidth,  valign=m]{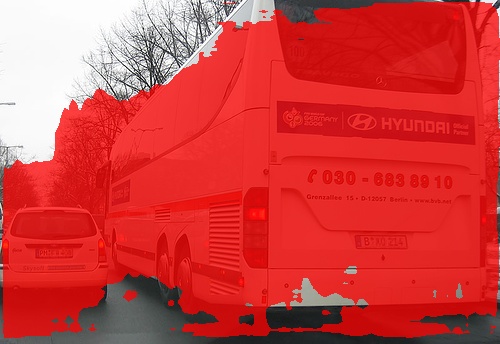} &
\includegraphics[width=0.065\linewidth,  valign=m]{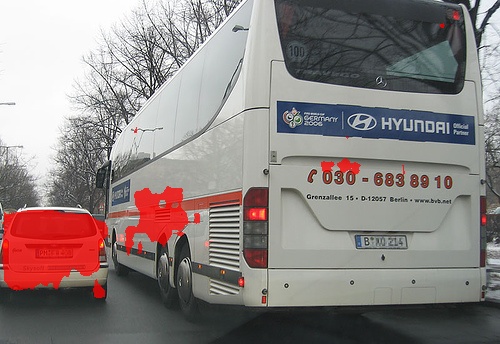} &
\includegraphics[width=0.065\linewidth,  valign=m]{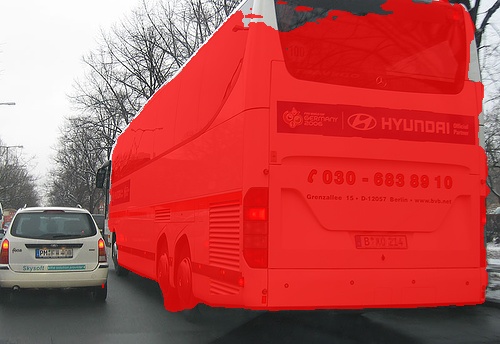} &
\includegraphics[width=0.065\linewidth,  valign=m]{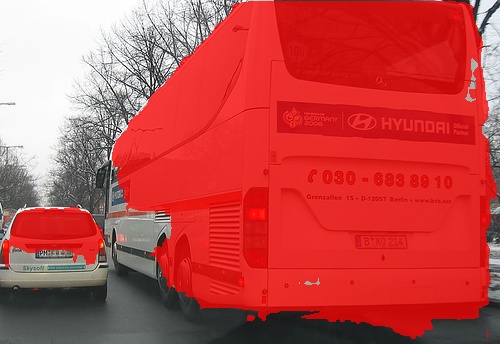} &
\includegraphics[width=0.065\linewidth,  valign=m]{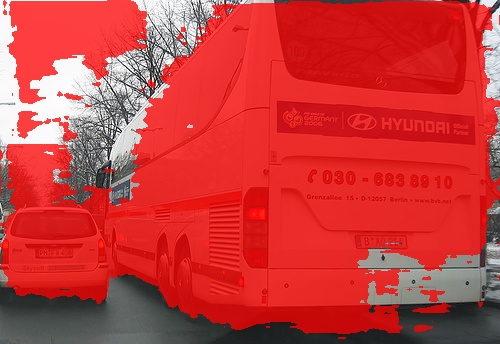} &
\includegraphics[width=0.065\linewidth,  valign=m]{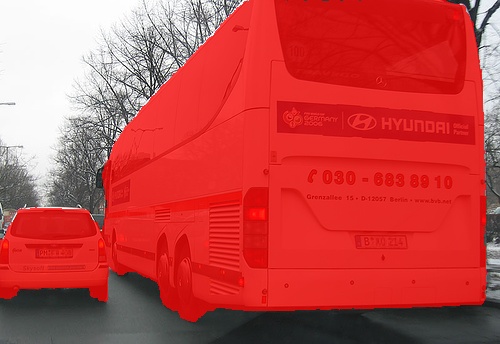} \\
& & 79.45 & 0.05 & 84.18 & 0.00 & 2.53 & 2.78 & 71.91 & 2.67 & 93.86 & 85.63 & 70.52 & 89.85		\\
\includegraphics[width=0.065\linewidth,  valign=m]{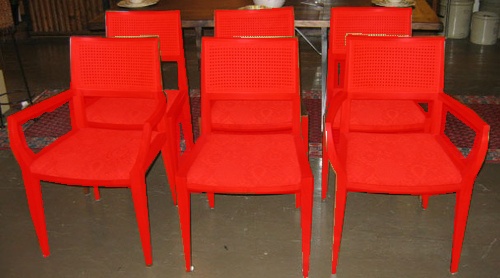} &
\includegraphics[width=0.065\linewidth,  valign=m]{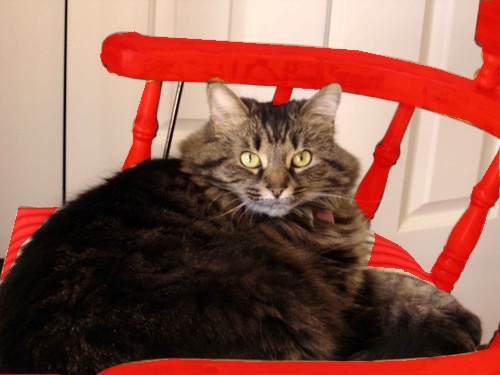} &
\includegraphics[width=0.065\linewidth,  valign=m]{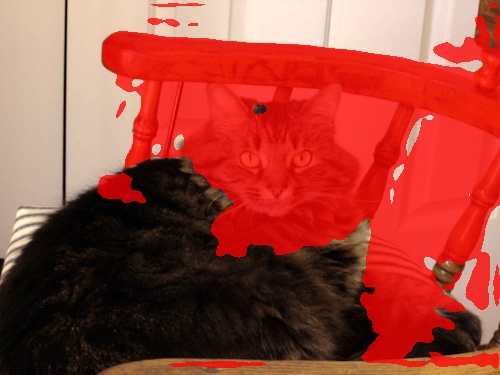} &
\includegraphics[width=0.065\linewidth,  valign=m]{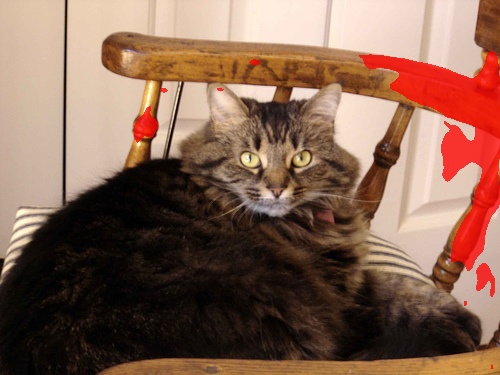} &
\includegraphics[width=0.065\linewidth,  valign=m]{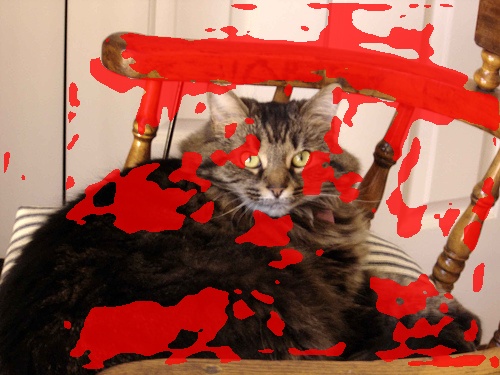} &
\includegraphics[width=0.065\linewidth,  valign=m]{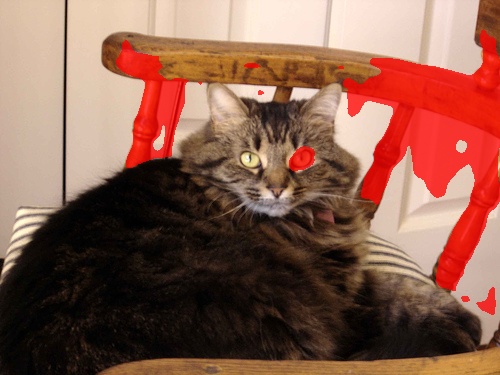} &
\includegraphics[width=0.065\linewidth,  valign=m]{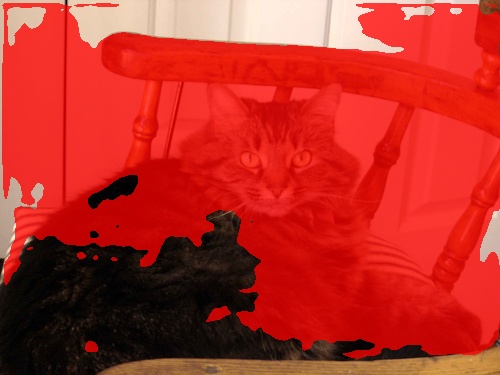} &
\includegraphics[width=0.065\linewidth,  valign=m]{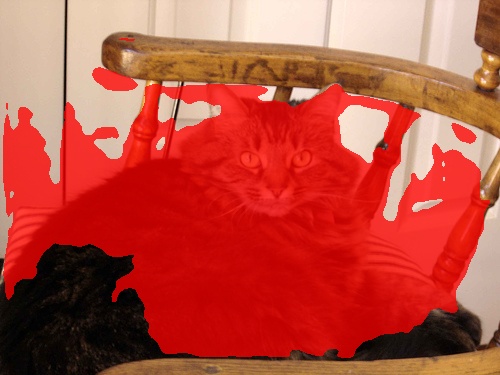} &
\includegraphics[width=0.065\linewidth,  valign=m]{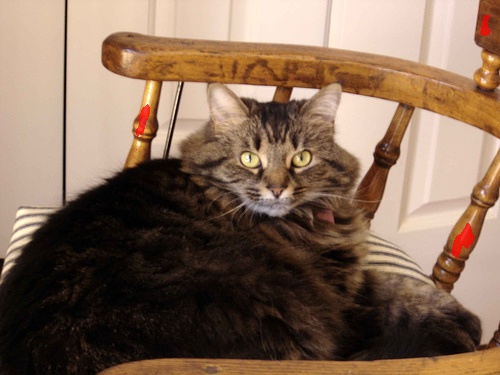} &
\includegraphics[width=0.065\linewidth,  valign=m]{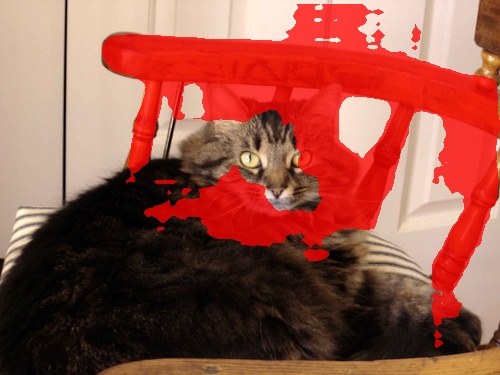} &
\includegraphics[width=0.065\linewidth,  valign=m]{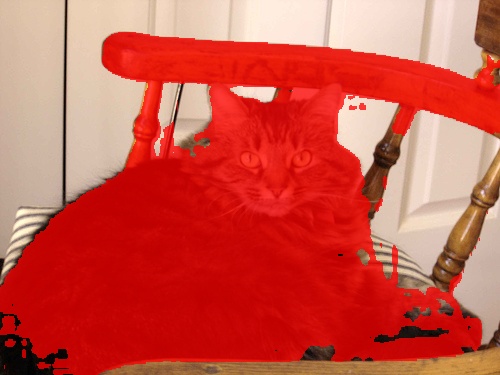} &
\includegraphics[width=0.065\linewidth,  valign=m]{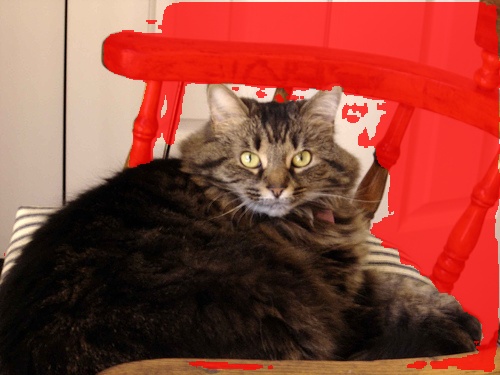} &
\includegraphics[width=0.065\linewidth,  valign=m]{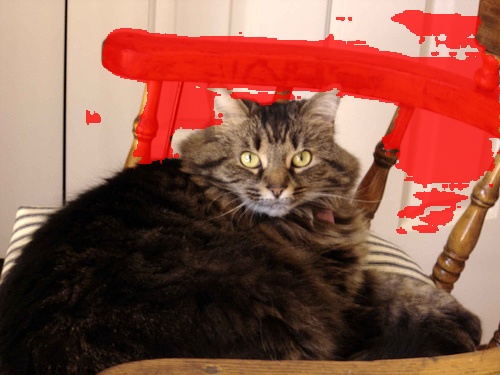} &
\includegraphics[width=0.065\linewidth,  valign=m]{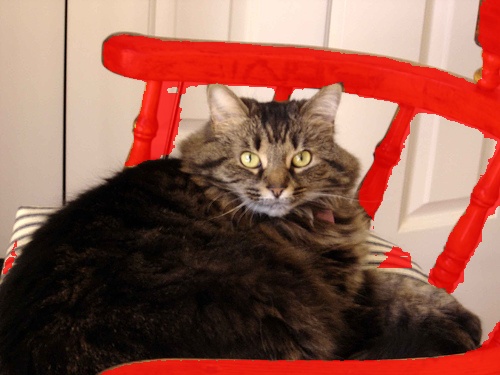} \\
& & 31.40 & 21.20 & 25.69 & 33.23 & 20.39 & 7.31 & 2.33 & 37.96 & 16.86 & 37.63 & 37.27 & 80.51		\\
\includegraphics[width=0.065\linewidth,  valign=m]{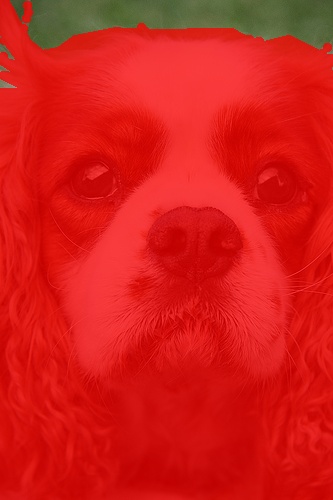} &
\includegraphics[width=0.065\linewidth,  valign=m]{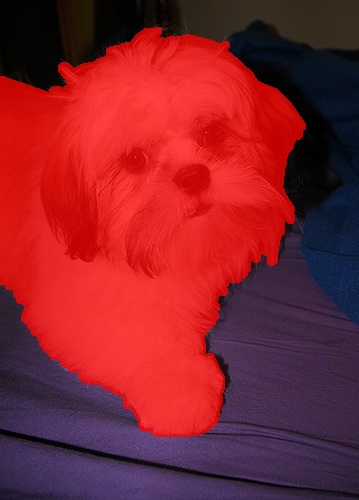} &
\includegraphics[width=0.065\linewidth,  valign=m]{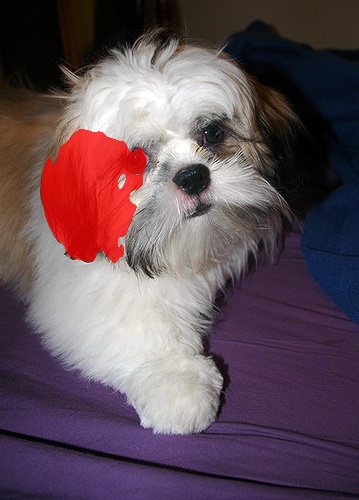} &
\includegraphics[width=0.065\linewidth,  valign=m]{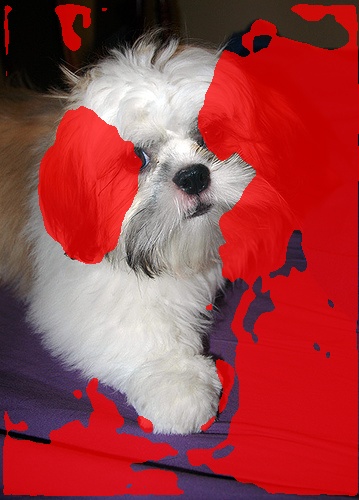} &
\includegraphics[width=0.065\linewidth,  valign=m]{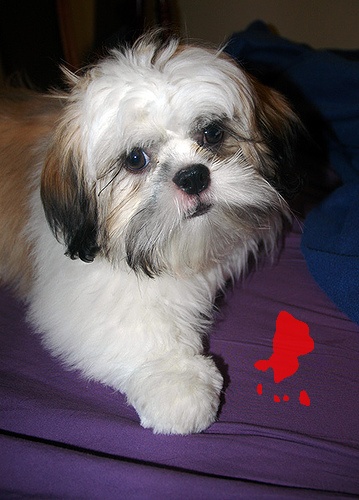} &
\includegraphics[width=0.065\linewidth,  valign=m]{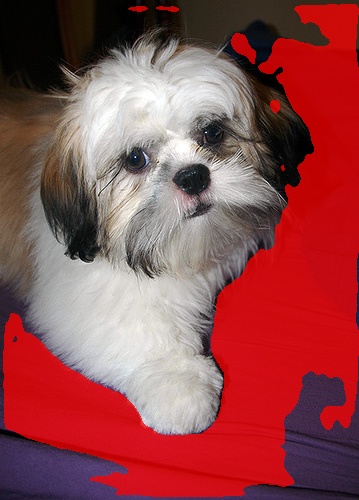} &
\includegraphics[width=0.065\linewidth,  valign=m]{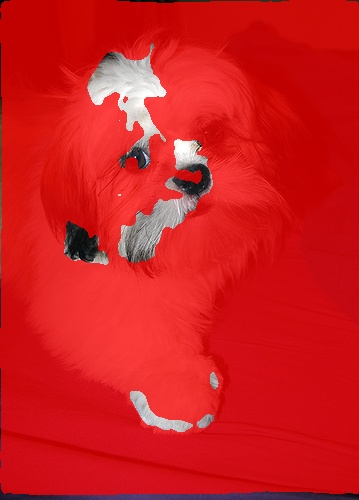} &
\includegraphics[width=0.065\linewidth,  valign=m]{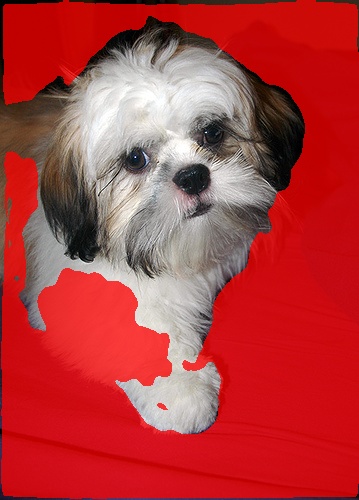} &
\includegraphics[width=0.065\linewidth,  valign=m]{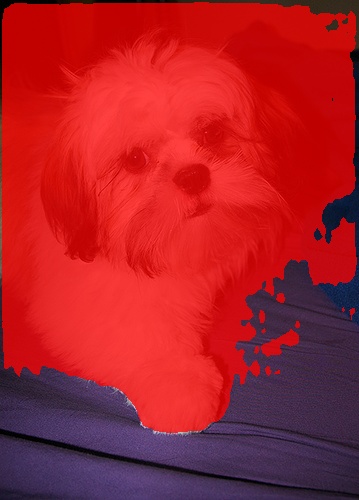} &
\includegraphics[width=0.065\linewidth,  valign=m]{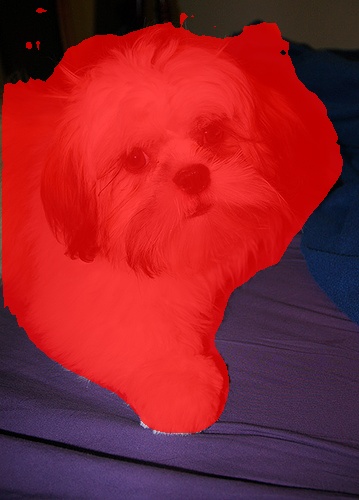} &
\includegraphics[width=0.065\linewidth,  valign=m]{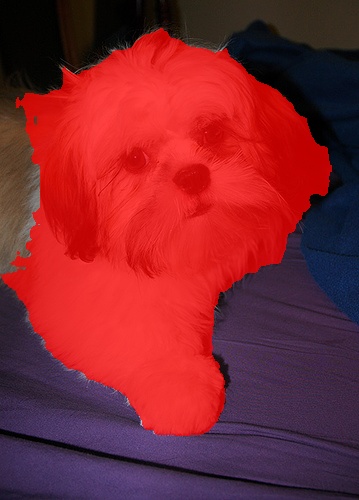} &
\includegraphics[width=0.065\linewidth,  valign=m]{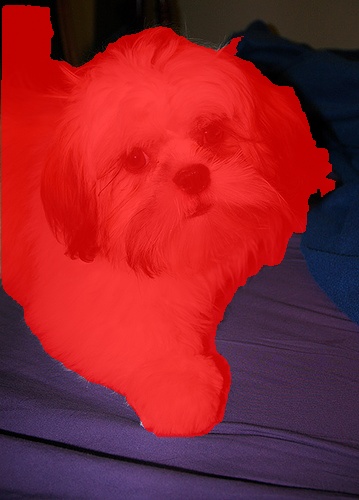} &
\includegraphics[width=0.065\linewidth,  valign=m]{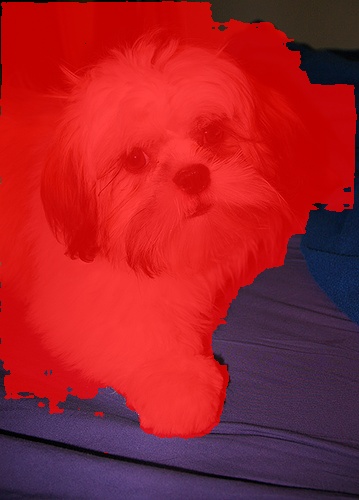} &
\includegraphics[width=0.065\linewidth,  valign=m]{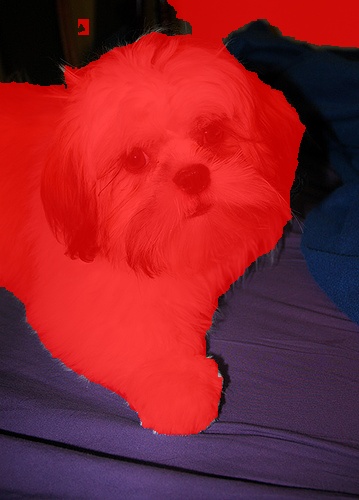} \\
& & 11.50 & 17.54 & 0.00 & 1.37 & 42.98 & 10.87 & 67.70 & 86.52 & 86.71 & 88.88 & 74.21 & 89.80		\\
\includegraphics[width=0.065\linewidth,  valign=m]{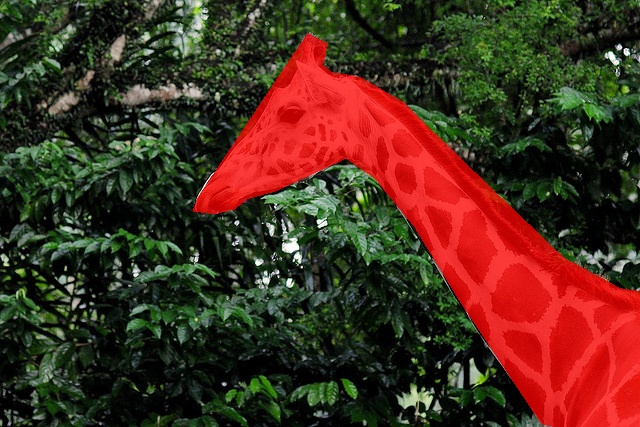} &
\includegraphics[width=0.065\linewidth,  valign=m]{qualitative_samples/query_set/giraffe.jpg} &
\includegraphics[width=0.065\linewidth,  valign=m]{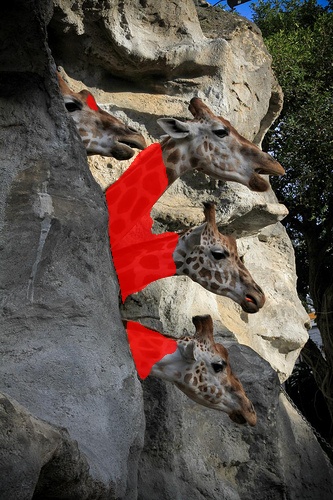} &
\includegraphics[width=0.065\linewidth,  valign=m]{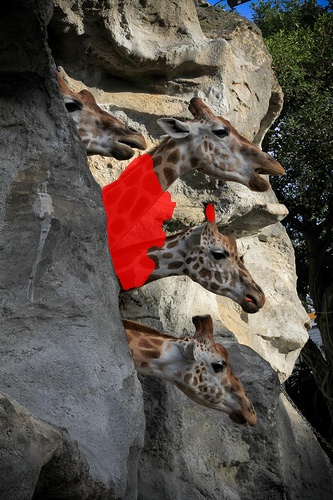} &
\includegraphics[width=0.065\linewidth,  valign=m]{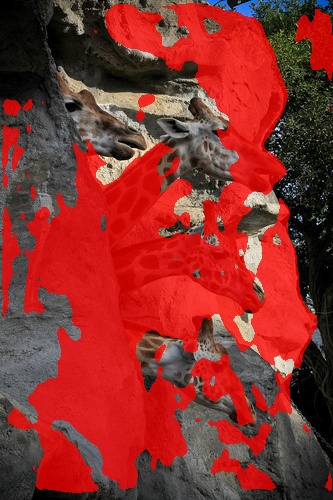} &
\includegraphics[width=0.065\linewidth,  valign=m]{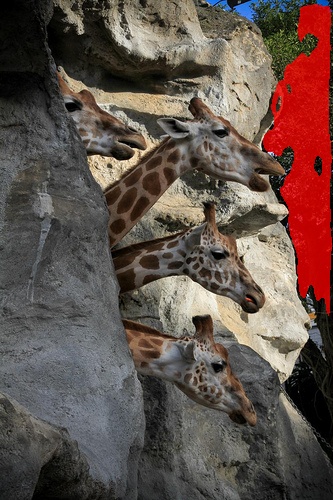} &
\includegraphics[width=0.065\linewidth,  valign=m]{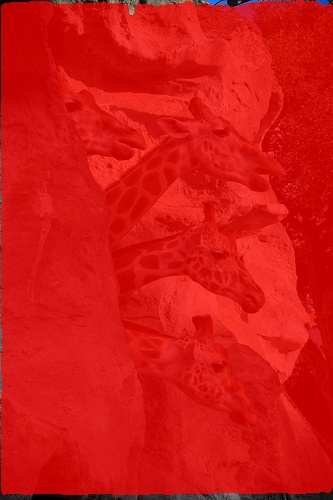} &
\includegraphics[width=0.065\linewidth,  valign=m]{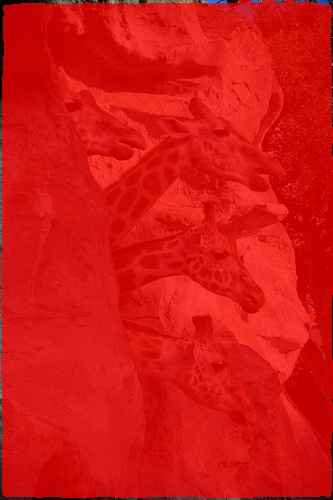} &
\includegraphics[width=0.065\linewidth,  valign=m]{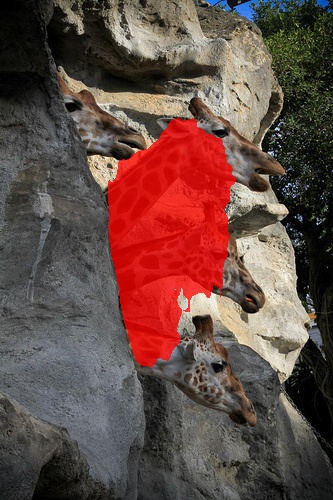} &
\includegraphics[width=0.065\linewidth,  valign=m]{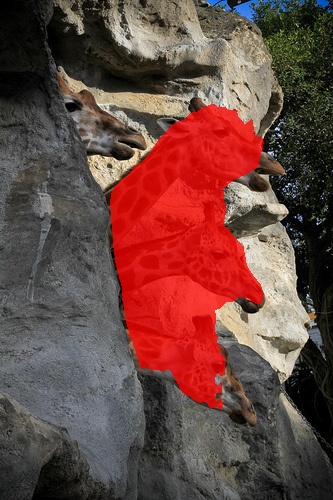} &
\includegraphics[width=0.065\linewidth,  valign=m]{qualitative_samples/VAT_res101_fss/giraffe.jpg} &
\includegraphics[width=0.065\linewidth,  valign=m]{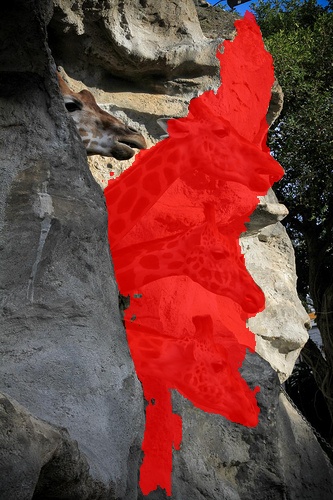} &
\includegraphics[width=0.065\linewidth,  valign=m]{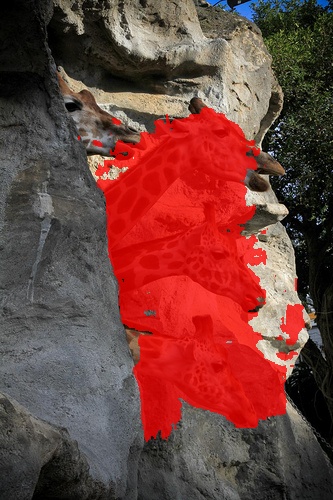} &
\includegraphics[width=0.065\linewidth,  valign=m]{qualitative_samples/SegGPT/giraffe.jpg} \\
& & 25.59 & 15.12 & 17.78 & 0.00 & 17.12 & 17.07 & 38.16 & 54.00 & 48.09 & 48.97 & 49.98 & 90.07	\\
\includegraphics[width=0.065\linewidth,  valign=m]{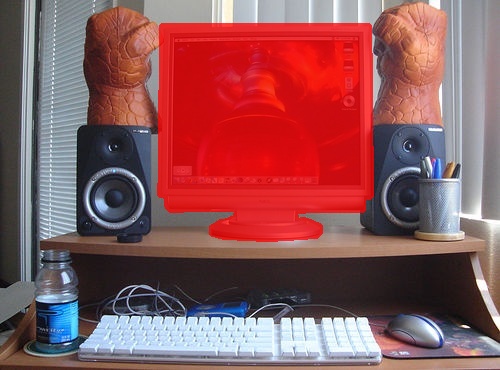} &
\includegraphics[width=0.065\linewidth,  valign=m]{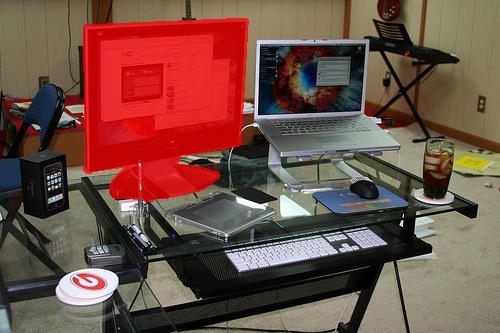} &
\includegraphics[width=0.065\linewidth,  valign=m]{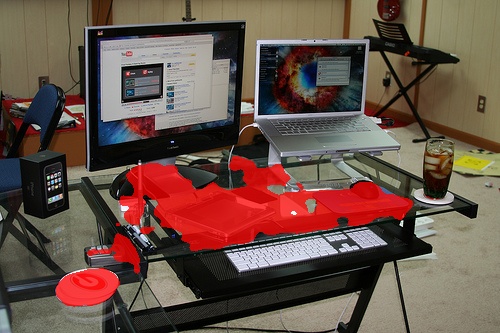} &
\includegraphics[width=0.065\linewidth,  valign=m]{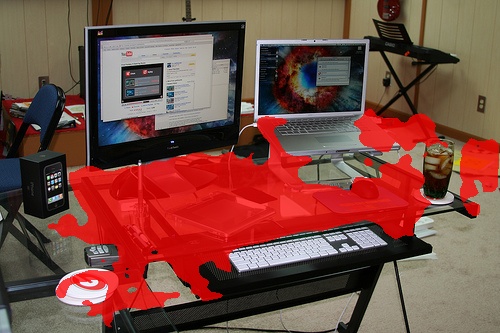} &
\includegraphics[width=0.065\linewidth,  valign=m]{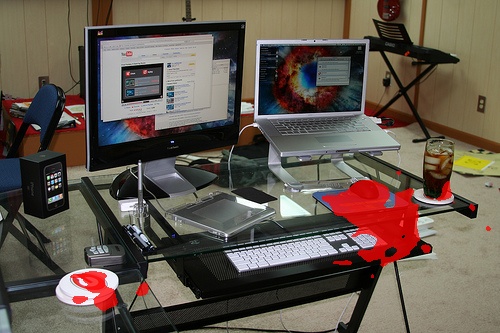} &
\includegraphics[width=0.065\linewidth,  valign=m]{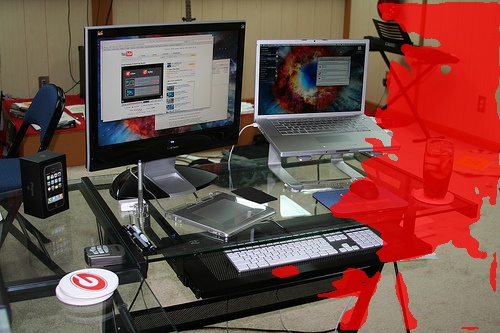} &
\includegraphics[width=0.065\linewidth,  valign=m]{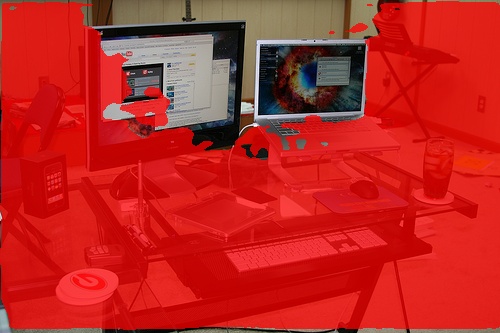} &
\includegraphics[width=0.065\linewidth,  valign=m]{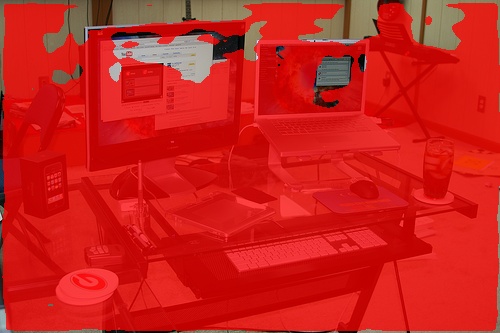} &
\includegraphics[width=0.065\linewidth,  valign=m]{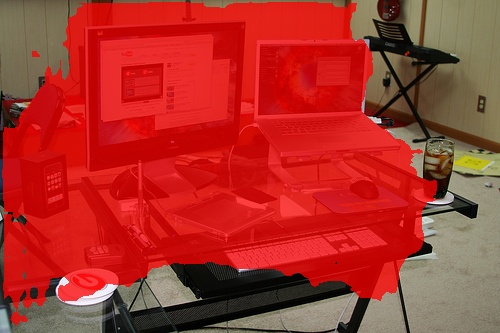} &
\includegraphics[width=0.065\linewidth,  valign=m]{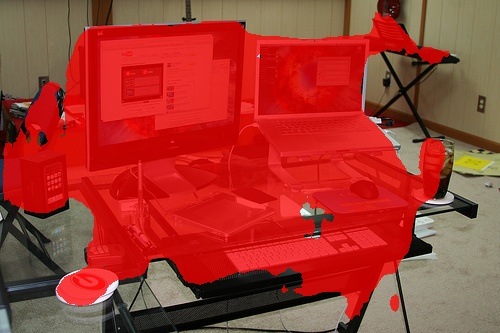} &
\includegraphics[width=0.065\linewidth,  valign=m]{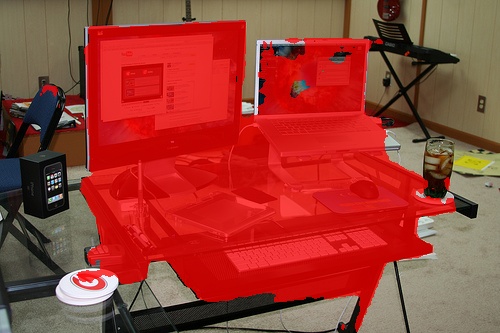} &
\includegraphics[width=0.065\linewidth,  valign=m]{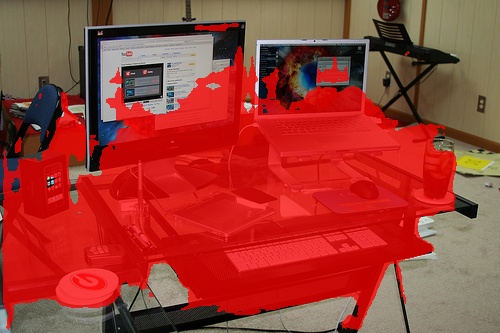} &
\includegraphics[width=0.065\linewidth,  valign=m]{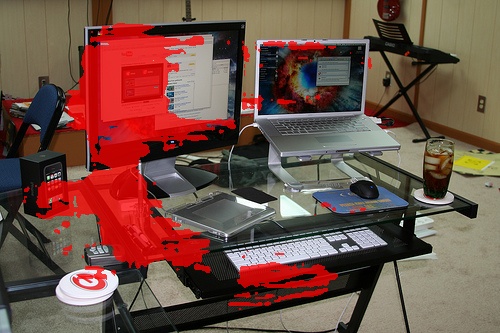} &
\includegraphics[width=0.065\linewidth,  valign=m]{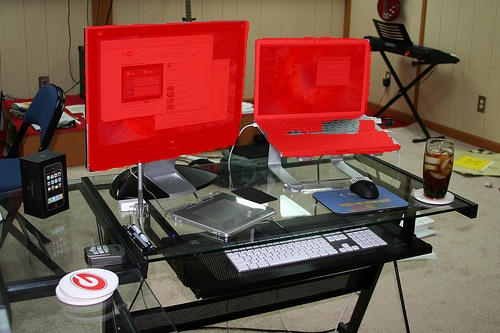} \\
& & 4.70 & 5.72 & 0.00 & 0.00 & 8.02 & 13.89 & 25.12 & 28.04 & 31.02 & 13.21 & 34.95 & 55.25		\\
\includegraphics[width=0.065\linewidth,  valign=m]{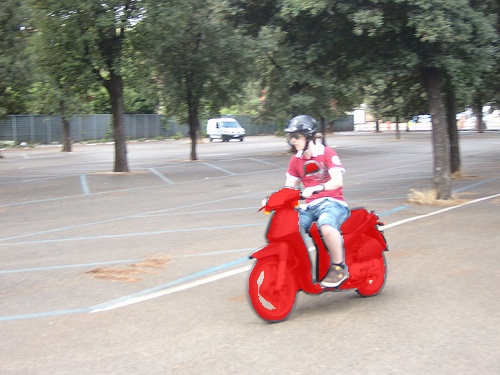} &
\includegraphics[width=0.065\linewidth,  valign=m]{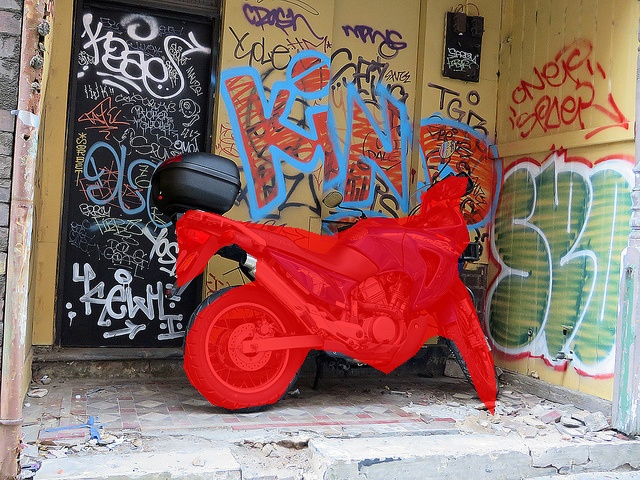} &
\includegraphics[width=0.065\linewidth,  valign=m]{qualitative_samples/DCAMA_resnet101_coco/motorbike.jpg} &
\includegraphics[width=0.065\linewidth,  valign=m]{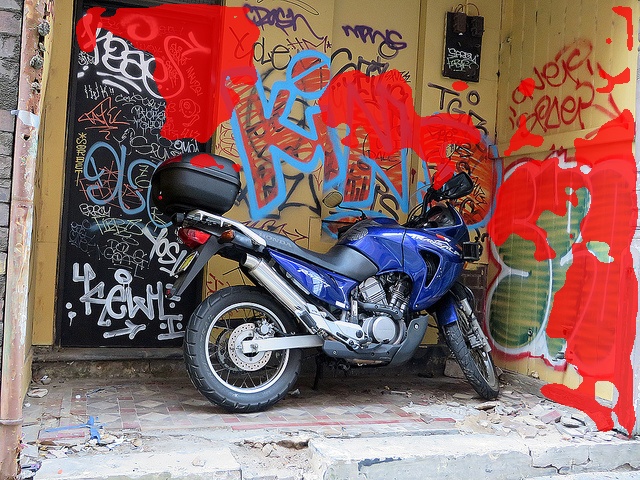} &
\includegraphics[width=0.065\linewidth,  valign=m]{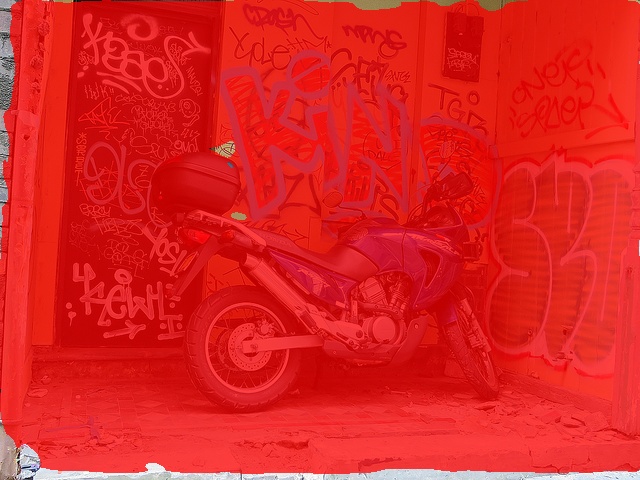} &
\includegraphics[width=0.065\linewidth,  valign=m]{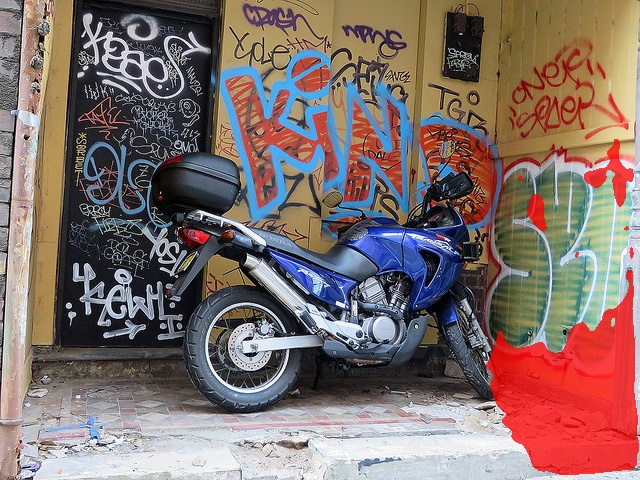} &
\includegraphics[width=0.065\linewidth,  valign=m]{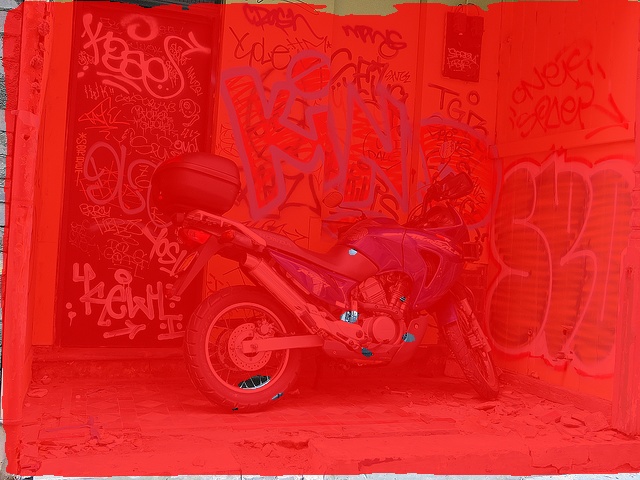} &
\includegraphics[width=0.065\linewidth,  valign=m]{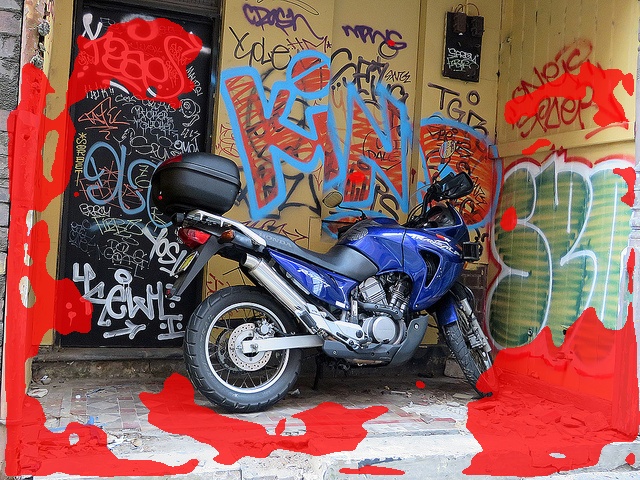} &
\includegraphics[width=0.065\linewidth,  valign=m]{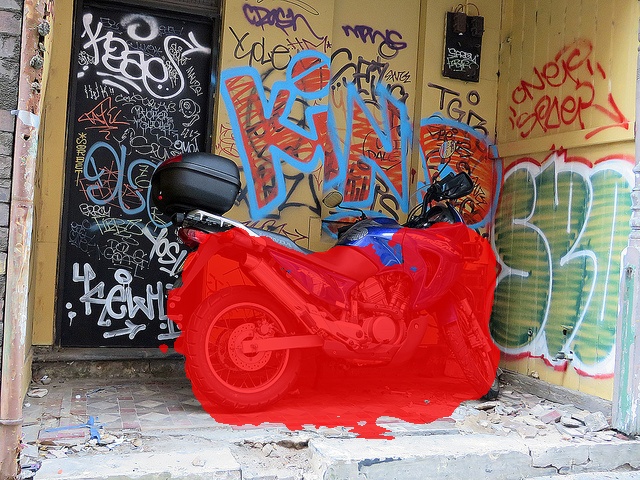} &
\includegraphics[width=0.065\linewidth,  valign=m]{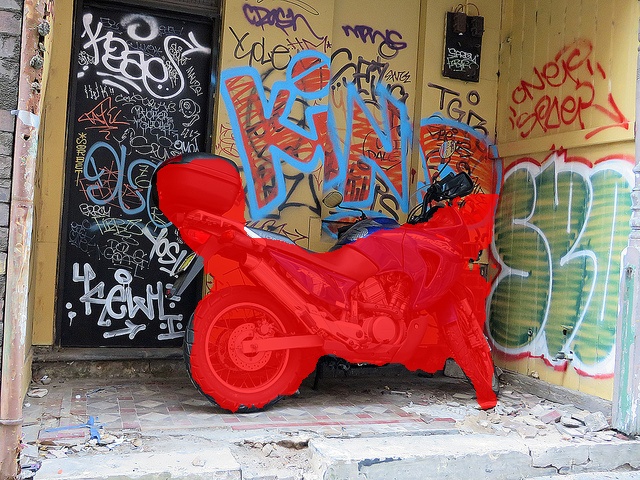} &
\includegraphics[width=0.065\linewidth,  valign=m]{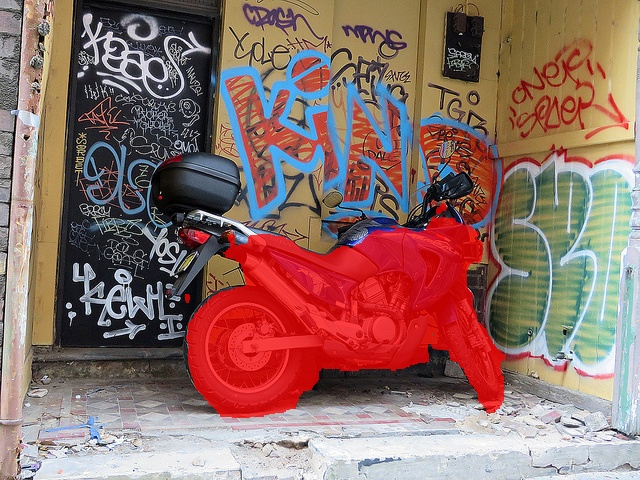} &
\includegraphics[width=0.065\linewidth,  valign=m]{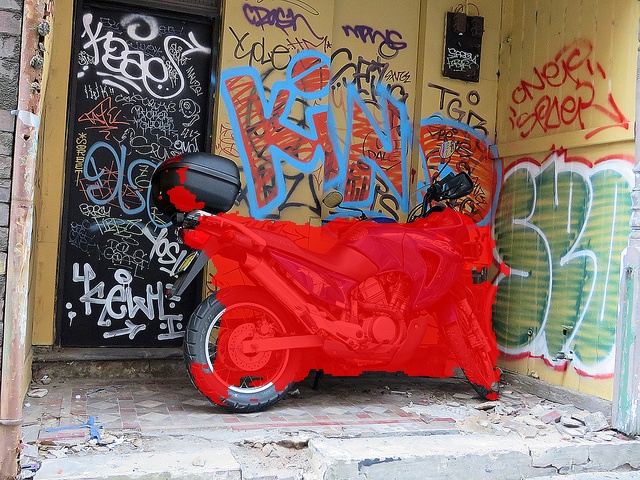} &
\includegraphics[width=0.065\linewidth,  valign=m]{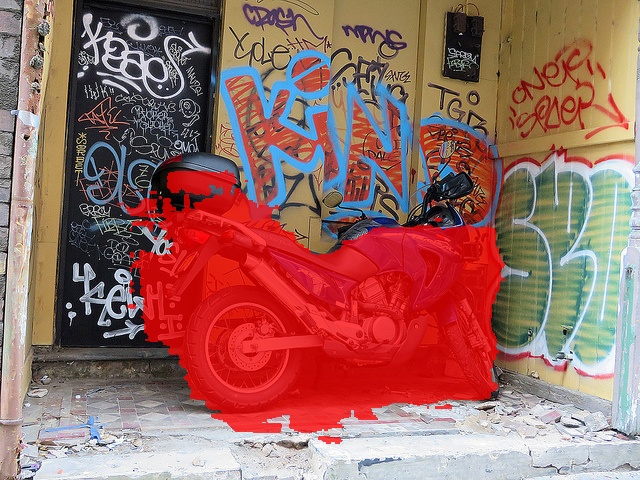} &
\includegraphics[width=0.065\linewidth,  valign=m]{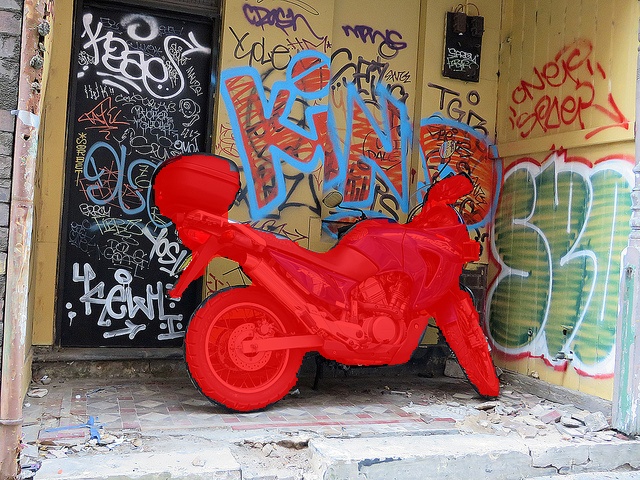} \\
& & 0.00 & 0.09 & 14.65 & 0.32 & 14.29 & 0.53 & 56.00 & 68.81 & 72.91 & 69.63 & 56.04 & 79.87		\\
\includegraphics[width=0.065\linewidth,  valign=m]{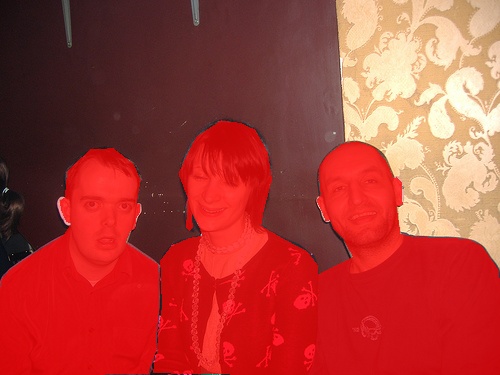} &
\includegraphics[width=0.065\linewidth,  valign=m]{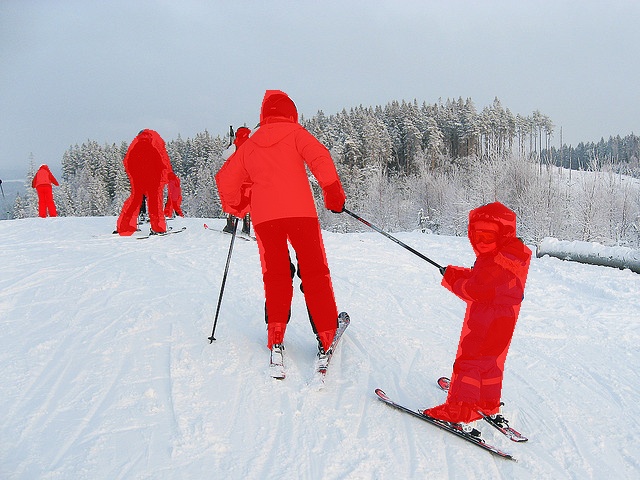} &
\includegraphics[width=0.065\linewidth,  valign=m]{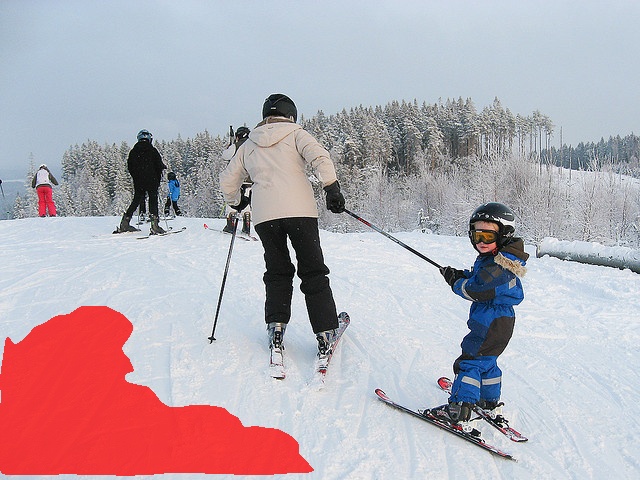} &
\includegraphics[width=0.065\linewidth,  valign=m]{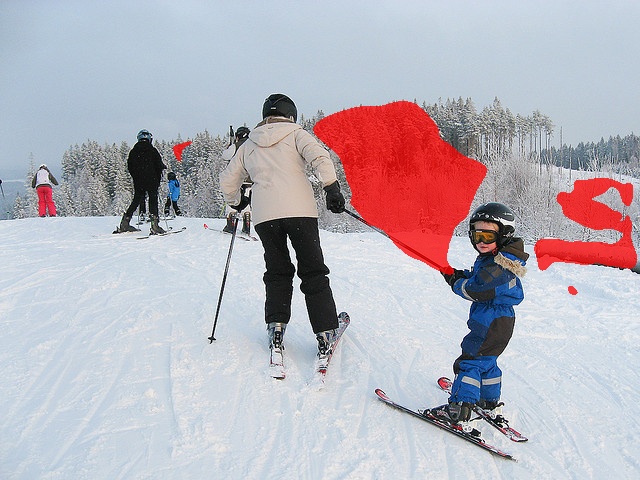} &
\includegraphics[width=0.065\linewidth,  valign=m]{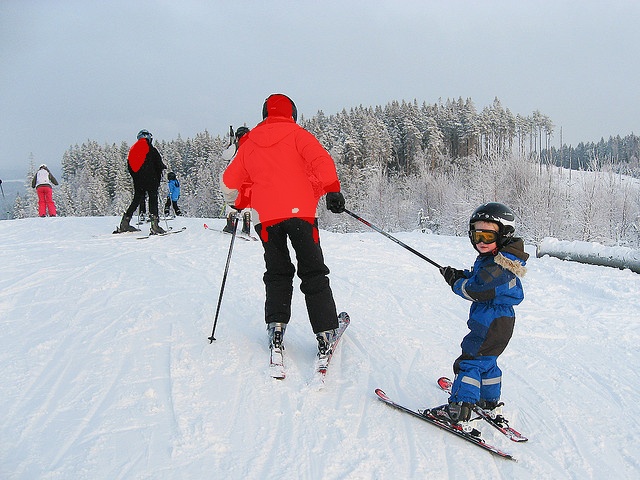} &
\includegraphics[width=0.065\linewidth,  valign=m]{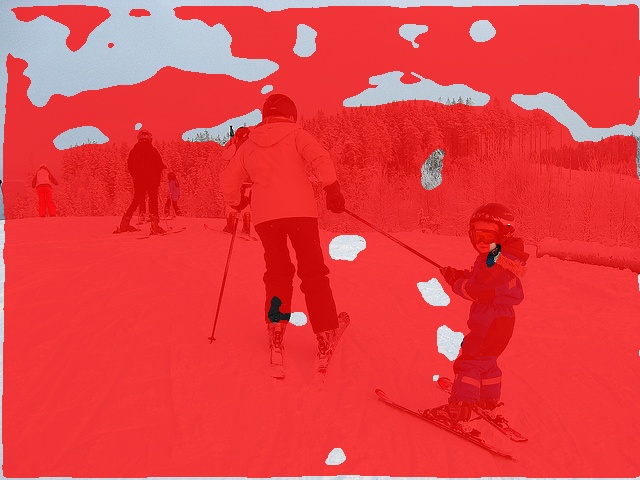} &
\includegraphics[width=0.065\linewidth,  valign=m]{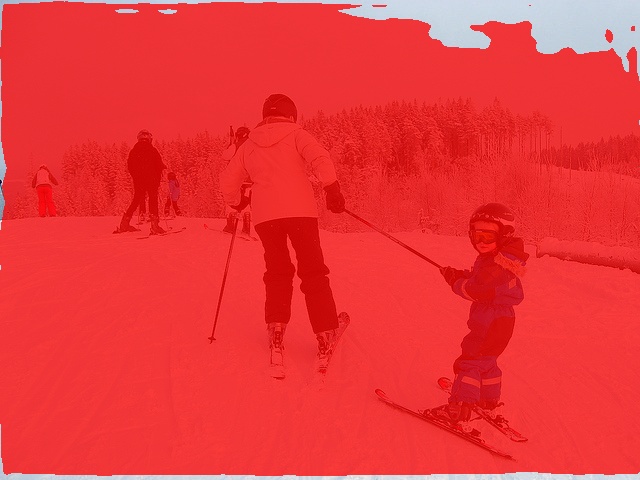} &
\includegraphics[width=0.065\linewidth,  valign=m]{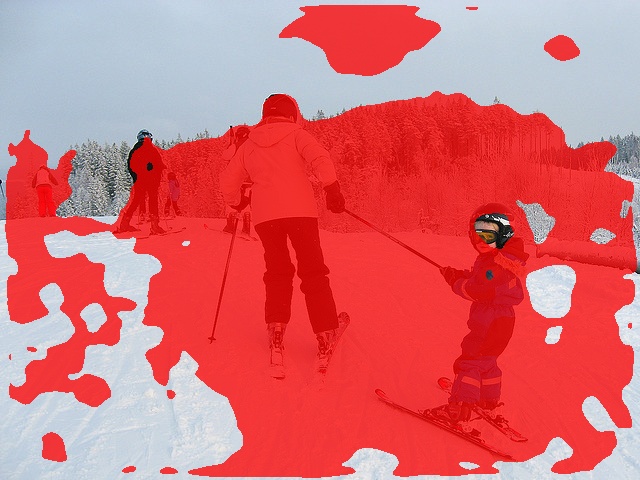} &
\includegraphics[width=0.065\linewidth,  valign=m]{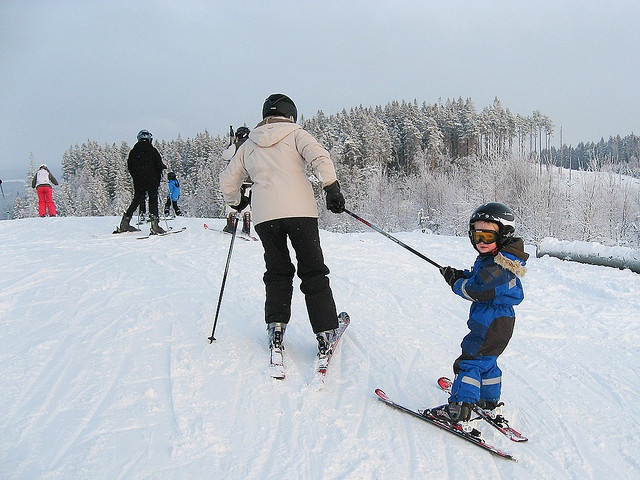} &
\includegraphics[width=0.065\linewidth,  valign=m]{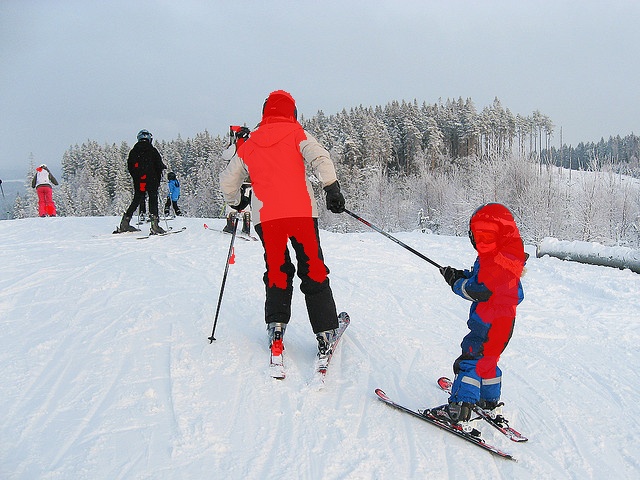} &
\includegraphics[width=0.065\linewidth,  valign=m]{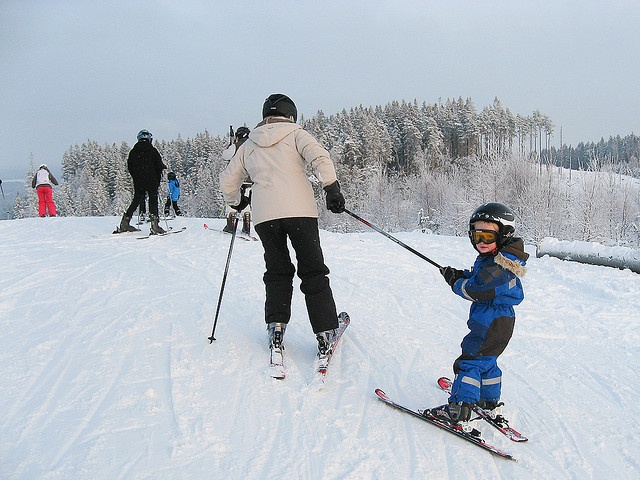} &
\includegraphics[width=0.065\linewidth,  valign=m]{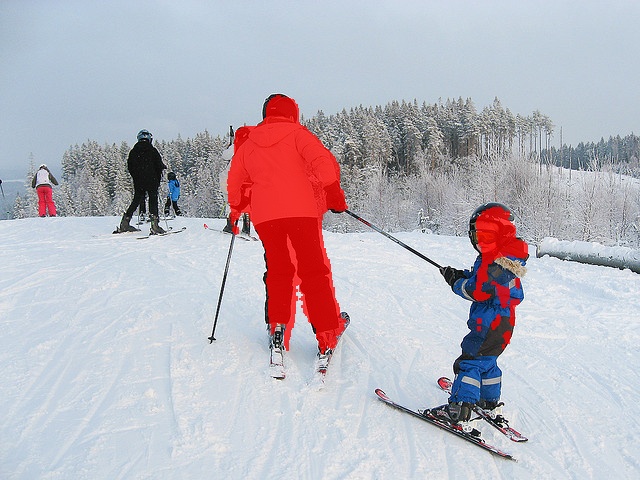} &
\includegraphics[width=0.065\linewidth,  valign=m]{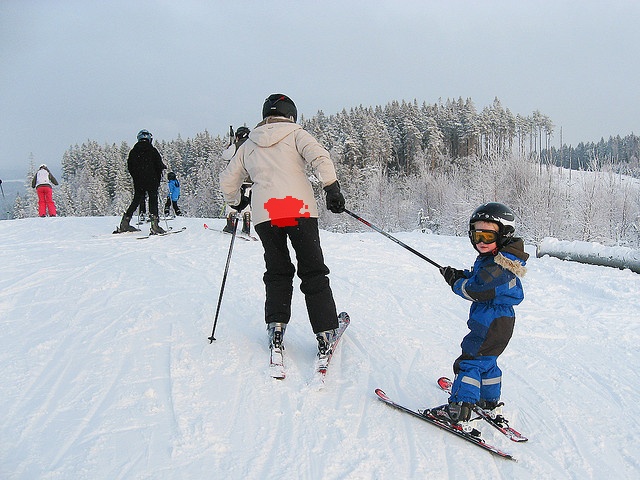} &
\includegraphics[width=0.065\linewidth,  valign=m]{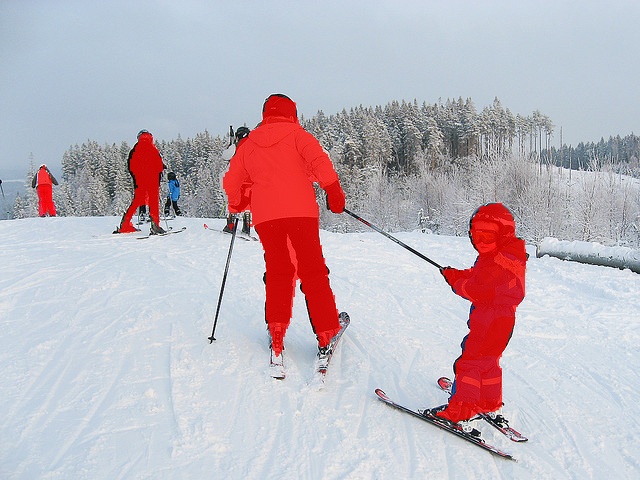} \\
& & 0.00 & 0.11 & 26.79 & 12.32 & 11.77 & 17.64 & 0.00 & 42.06 & 0.00 & 51.01 & 2.74 & 80.11		\\

\end{tabular}

\label{tab_qualitative_samples}
\end{table} 

\section{Outlook in Few Shot Segmentation}
In this section, we are going to highlight the limitations and criticisms of the \acrshort{fss} research line and models. We are also going to showcase proposed extensions of the presented \acrshort{fss} problem that could benefit other research fields.

\subsection*{Limitations and open challenges of episodic training }
Episodic training, which is widely used and accepted as a good approach for  \acrshort{fss}, has some drawbacks as noted by Cao et al. in  \cite{cao2019theoretical}. Indeed, in real-world applications it can be difficult to predict and fix the number of available shots $K$ as it ranges between applications. Cao et al. also theoretically demonstrated how this mismatch between the value of $K$ during meta-training and meta-testing can lead to a decline in performance. Additionally, they showed that a model based on prototypes does not significantly improve its prediction accuracy by simply increasing the number of labelled samples in the support set. 

Given these limitations, Boudiaf et al. \cite{boudiaf2021few} proposed the use of transductive learning. Unlike inductive learning, transductive learning techniques have access to all data, both training and testing, beforehand. The key idea of this approach is that during the learning process it is possible to make use of the patterns and additional information present in the testing data, even without knowing the labels. In such a way it could be possible to better exploit the few annotated examples typical of \acrshort{fss}.

\subsection*{N-Way K-Shot}

A great difference between \acrshort{fss} and traditional supervised semantic segmentation is that  \acrshort{fss} models learn to segment only one class, while most modern semantic segmentation models are capable of segmenting many different classes. To address this limitation, Dong et al. in \cite{Dong2018FewShotSS} generalized the \acrshort{fss} problem to $N$ classes naming it N-Way K-Shot Semantic Segmentation. 
In this setting we can define \acrfull{nway_kshot_fss}  as follow:
\begin{definition}[N-Way K-Shot Semantic Segmentation]
 N-Way K-Shot Semantic Segmentation is the problem of predicting the semantic segmentation mask $\hat{M}_q$ of the $N$ different classes of the label-set $L$ in a query image $I_q$ given a support set $S$  and $K$ examples for each class.
\end{definition}

\noindent
More formally, defined a label set $L$ of $N$ classes:
\[
    L = \{l_i\}_{i=1}^N
\]
we can build a support set:
\[
S = \{(I^j, Y^j_{l_i})\}_{j=1,i=1}^{j=K,i=N}
\]
where is allowed $I^{i} = I^{j}$ as long as $Y^{i} \neq Y^{j}$. Is worth pointing out that in N-Way K-Shot \acrshort{fss} more than one class subject can be present in a given image. 
In addition, in this extension of \acrshort{fss}, also the output of the model is changed. Since the model is segmenting $N$ classes plus the background, given a query image $I_q$ with resolution $[H^q, W^q]$, the output of the model will  have the shape of $[H^q, W^q, N+1]$ .

\subsection*{Generalized Few Shot Segmentation}
Tian et al. in \cite{tian2022Generalized} highlight the limitations of traditional \acrshort{fss} models, which can only predict novel classes present in the support set and cannot predict base classes used for pretraining. They also stress that support and query images must contain the same classes for the model to perform optimally, meaning that if the query image does not contain the subject of the support set, the model may not perform optimally and could potentially misbehave. In other words, the model expects to find the same objects in both the support set and query image, but this is not always guaranteed in real-life situations.

To address these limitations, Tian et al. in \cite{tian2022Generalized} and Lu et al. \cite{10145054} envision a segmentation model that can both segment base classes learned from a sufficient number of samples and learn new classes in a few-shot manner. This problem is defined as \acrfull{GFS-Seg} in their work.
A potential difficulty of this setting is that the unbalanced number of training examples for the base and novel classes could lead to biases in the segmentation model. On this aspect, Liu et al. \cite{Liu2023HarmonizingBA} speculate that as the base classes are learned with a higher number of examples, and are thus better learned, their prototypes should not be updated as much as the prototypes for novel classes which are learned from scratch. Therefore, they propose discouraging large modifications to base class prototypes while boosting large distances between prototypes with a class contrastive loss and a class relationship loss. 
With a similar intent, the Project onto Orthogonal Prototype (POP) \cite{10204345} framework builds a set of orthogonal prototypes where the prototypes relative to the base classes remain frozen during meta training.

Aiming at a more real-world-oriented model, Hajimiri et al. \cite{10204373} propose to extend the concept introduced by Tian et al. \cite{tian2022Generalized}. Starting from \acrshort{GFS-Seg}, they incorporate the capability to predict a multi class segmentation mask, and not just a binary mask typical of many works in \acrshort{fss}. Their approach allows for the prediction of multiple classes on the query image, including both the base classes learned during pretraining and the novel classes learned in few-shot mode. Effectively bridging the gap between \acrshort{GFS-Seg} and N-Way K-Shot Semantic Segmentation.

Lei et al. \cite{cdfss} identify yet another crucial gap in the deployment of Few-Shot Segmentation (FSS) models for real-world applications, notably the inherent assumption that base and novel classes originate from the same dataset domain, typically PASCAL-5\textsuperscript{i} or COCO-20\textsuperscript{i}.

However, this assumption fails to hold in practical scenarios, prompting the extension of \acrshort{fss} into \acrfull{cdfss}. \acrshort{cdfss} aims at crafting more versatile models capable of addressing diverse domain gaps. 
To this end, Lei et al. introduce a benchmark dataset comprising four distinct domains: FSS-100 \cite{li2020fss} for everyday object images, Deepglobe \cite{demir2018deepglobe} offering satellite images, ISIC2018 \cite{Codella2019SkinLA, Tschandl2018TheHD} featuring dermoscopic images of skin lesions, and the Chest X-Ray dataset \cite{Candemir2014LungSI,Jaeger2014AutomaticTS} containing X-ray images. 

This diverse dataset facilitates the implementation of \acrshort{cdfss} training protocol, involving the division of the dataset into $D_{train}$ and $D_{test}$ to sample training episode and test episode, respectively. Crucially, $D_{train}$ and $D_{test}$ are designed to not have any overlap in terms of both image content and segmented classes, ensuring a substantial domain gap. The choice of subsets for $D_{train}$ and $D_{test}$ introduces varying degrees of domain shift, quantified by the authors with  \acrfull{fid} \cite{Heusel2017GANsTB}. 

This extension of FSS has spurred the development of robust models resilient to domain gaps, with subsequent works \cite{cdfss, Huang_2023_BMVC, Fan2023DARNetBD, Chen_2024_WACV} emphasizing the importance of translating domain-specific learned features into domain-agnostic representations while preserving the ability to discern target-specific features.

\subsection*{Incremental Few Shot Segmentation}
With similar premises of \acrshort{GFS-Seg}, Siam et al. in \cite{siam2019adaptive} and Carmelli et al. in \cite{cermelli2020prototype} proposes to extend the \acrshort{fss} problem by setting it as incremental learning. Their rationale is that in applications such as robotics, once a segmentation model is deployed it might over time need to learn new classes while keeping the previously learned. Cermelli et al. named this problem \acrfull{ifss}.
The main difference with \acrshort{GFS-Seg} is that in \acrshort{ifss} the base dataset used to pretrain the model on base classes is no longer available to the model, which can access only the data of the last episode to learn new classes and update the older. More recently Zhou et al. \cite{aifss} highlight that in \acrshort{ifss} semantic confusion between base and novel classes might diminishes segmentation accuracy. They propose a method for semantic-guided relation alignment and adaptation to address this issue. Initially, they align base class representations with their semantic information, then conduct semantic-guided adaptation during incremental learning to ensure consistency between visual and semantic embeddings of encountered novel classes, thus reducing semantic overlap.

To simulate such a scenario Siam et al. used the PASCAL dataset in an incremental manner, naming it iPASCAL, to provide the model with two new classes per episode. The model they developed for this task is based on prototypical learning with a key difference: it keeps a rolling average for the known class prototypes and makes new class prototypes for the new classes. Using this approach their model is capable of keeping and updating the internal representation of the base classes while also being able of learning new ones.

\section{Conclusions}
Semantic Segmentation is a crucial computer vision task that has a wide range of applications, from healthcare to robotics and surveillance. However, a major challenge in applying state-of-the-art models to these fields is the requirement for a large labelled training dataset. Collecting and labelling such an amount of data is both expensive and time-consuming, and it can also raise privacy concerns as well as technical issues.
Recently, the field of \acrfull{fsl} has emerged to tackle the problem of learning a new task with limited training samples. This paper aims at  examining the current challenges in applying \acrshort{fsl} to the task of semantic segmentation, known as \acrfull{fss},  and at providing a general overview of the main approaches  found in the literature of this topic. Most works in this area can be categorised  based  on the main strategy they adopt to solve the task. We identified the major approaches to be Conditional Networks, Prototypical Networks, and Latent Space Optimization.

Conditional Networks were the first solution deployed for \acrshort{fss},  the challenge with these models is determining the best parameter set for the conditioning branch to pass to the segmentation branch. The parameter set must be informative enough to recognize the target class in the query image while allowing for appearance variation.

Prototypical Networks use a module to compute class representative prototypes, which serve as references to evaluate if a pixel in the query image belongs to a certain class. The main challenges with this approach are the metric used to compute the similarity between a pixel and the class prototype, as well as how to compute the prototype in order to make it informative enough for describing a whole class.

Latent Space Optimization approaches study how a well-structured latent space can aid in \acrshort{fss}. Some works train a GAN on the target class to produce a full-sized dataset, while others describe classes as statistical distributions in the latent space. The main limitation of this approach is that results are heavily dependent on the difficult training procedures of GANs.

Furthermore, we delineate how contemporary foundational models can be tailored for the \acrfull{fss}, elucidating the potential of prompt engineering and multimodal models to pave the way for novel research trajectories. Additionally, we underscore the significance of generalist models, which possess the capacity to address diverse tasks, including the \acrfull{fss}.

 We also present a comprehensive overview of benchmark datasets, metrics, and significant results in the field of \acrshort{fss} summarizing the results of the most important works in this area, while also presenting a selection of qualitative results for the task of One Shot Semantic Segmentation.

Finally, this paper addresses  some of the challenges and potential developments in \acrshort{fss}. Some researchers have argued that the traditional approach of episodic training may not be optimal, as mismatches between the number of shots available for meta-training and meta-testing can result in significant performance degradation. We explore N-Way K-Shot Semantic Segmentation as the task of learning to segment multiple classes with limited shots per class. Other authors have emphasized the importance of not forgetting the classes used during pretraining, and the need to continuously learn new classes, which has led to the development of \acrfull{GFS-Seg} and \acrfull{ifss} methods.

In conclusion, the field of \acrshort{fss} is still open and active, and we can expect to see new and consequential works in the upcoming years. Ongoing developments and new approaches are expected to address the challenges in this area enabling the use of semantic segmentation models in new fields.

\begin{acks}
This paper has been partially supported by “Agritech: Centro Nazionale per lo sviluppo delle nuove tecnologie in agricoltura" project funded by the European Union NextGenerationEU program within the PNRR.
\end{acks}

\bibliographystyle{ACM-Reference-Format}
\bibliography{refs.bib}

\end{document}